\documentclass[10pt,twocolumn,letterpaper]{article}
\usepackage{geometry}
\usepackage{cvpr}
\usepackage{times}
\usepackage{epsfig}
\usepackage{graphicx}
\usepackage{amsmath}
\usepackage{amssymb}
\usepackage{gensymb}

\usepackage[linesnumbered, ruled]{algorithm2e}
\usepackage{algorithmic}
\usepackage[utf8]{inputenc} % allow utf-8 input
\usepackage[T1]{fontenc}    % use 8-bit T1 fonts
\usepackage{url}            % simple URL typesetting
\usepackage{booktabs}       % professional-quality tables
\usepackage{amsfonts}       % blackboard math symbols
\usepackage{nicefrac}       % compact symbols for 1/2, etc.
\usepackage{microtype}      % microtypography
\usepackage{amsmath,amssymb}
\usepackage{color}
\usepackage{float}
\usepackage{array}
\usepackage{multirow}
\usepackage{booktabs}
\usepackage{subfigure}
\usepackage{amsthm}
\usepackage{wrapfig,lipsum,booktabs}
\usepackage{xcolor}
\usepackage{bm}
\usepackage{textcomp}

\newcommand{\Koven}[1]{{\color{black}#1}}
\newcommand{\K}[1]{{\color{black}#1}}
\newcommand{\kk}[1]{{\color{black}#1}}
\newcommand{\kkk}[1]{{\color{black}#1}}
\newcommand{\ws}[1]{{\color{black}#1}}

\newcommand{\bigfont}{\fontsize{21pt}{\baselineskip}\selectfont}

\usepackage[breaklinks=true,bookmarks=false]{hyperref}

\cvprfinalcopy % *** Uncomment this line for the final submission

\def\cvprPaperID{2238} % *** Enter the CVPR Paper ID here

\ifcvprfinal\fi
\begin{document}

	{\onecolumn
		
		\noindent \textbf{\bigfont{Weakly supervised discriminative feature learning with state information for person identification}}
		
		\vspace{2cm}
		
		\noindent {\LARGE{Hong-Xing Yu, Wei-Shi Zheng}}
		
		%\Large
		\vspace{2cm}
		
		\large{
		\noindent Code is available at: \\
		\ \ \ \ \ \ \ \ \ \ \ \ \url{https://github.com/KovenYu/state-information}
		
		\vspace{1cm}
		
		\noindent For reference of this work, please cite:
		
		\vspace{1cm}
		\noindent Hong-Xing Yu and Wei-Shi Zheng.
		``Weakly supervised discriminative feature learning with state information for person identification''
		In \emph{Proceedings of the IEEE International Conference on Computer Vision and Pattern Recognition (CVPR).} 2020.
		
		\vspace{1cm}
		
		\noindent Bib:\\
		\noindent
		@inproceedings\{yu2020weakly,\\
		\ \ \   title=\{Weakly supervised discriminative feature learning with state information for person identification\},\\
		\ \ \  author=\{Yu, Hong-Xing and Zheng, Wei-Shi\},\\
		\ \ \  booktitle=\{Proceedings of the IEEE International Conference on Computer Vision and Pattern Recognition (CVPR)\},\\
		\ \ \  year=\{2020\}\\
		\}
	}
}
	
	%\clearpage
	%
	%\newpage
	\restoregeometry

%%%%%%%%% TITLE
\title{Weakly Supervised Discriminative Feature Learning with State Information \\for Person Identification}

\author{Hong-Xing Yu\\
Sun Yat-sen University, China\\
{\tt\small xkoven@gmail.com}
% For a paper whose authors are all at the same institution,
% omit the following lines up until the closing ``}''.
% Additional authors and addresses can be added with ``\and'',
% just like the second author.
% To save space, use either the email address or home page, not both
\and
Wei-Shi Zheng\\
Sun Yat-sen University, China\\
{\tt\small wszheng@ieee.org}
}

\maketitle
\thispagestyle{empty}

\begin{abstract}
\vspace{-0.2cm}
Unsupervised learning of identity-discriminative visual feature
is appealing in real-world tasks where manual labelling is costly.
However, the images of an identity
can be visually discrepant when images are taken under different \emph{states}, e.g. different camera views and poses.
This visual discrepancy leads to great difficulty in unsupervised discriminative learning.
Fortunately, in real-world tasks we could often know the states without human annotation,
e.g. we can easily have the camera view labels in person re-identification and facial pose labels in face recognition.
In this work we propose utilizing the state information as weak supervision to address the visual discrepancy caused by different states.
We formulate a simple pseudo label model
and utilize the state information in an attempt to refine the assigned pseudo labels
by the weakly supervised decision boundary rectification and weakly supervised feature drift regularization.
We evaluate our model on unsupervised person re-identification and pose-invariant face recognition.
Despite the simplicity of our method,
it could outperform the state-of-the-art results on Duke-reID, MultiPIE and CFP datasets with a standard ResNet-50 backbone.
We also find our model could perform comparably with the standard supervised fine-tuning results on the three datasets. Code is available at \url{https://github.com/KovenYu/state-information}.
\end{abstract}

\section{Introduction}
While deep discriminative feature learning has shown great success in many vision tasks,
it depends highly on the manually labelled large-scale visual data.
This limits its scalability to real-world tasks where the labelling is costly and tedious,
e.g. person re-identification \cite{2016_Arxiv_Zheng,2015_TCSVT_xiaojuan} \kk{and unconstrained pose-invariant face recognition \cite{zhao2018towards}.}
Thus, learning identity-discriminative features without manual labels has drawn increasing attention 
due to its promise to address the scalability problem \cite{2017_ICCV_asymmetric,2017_Arxiv_PUL,2019_CVPR_MAR,2019_TPAMI_DECAMEL,2019_CVPR_patch}.

\begin{figure}[t]
\begin{center}
\includegraphics[width=0.9\linewidth]{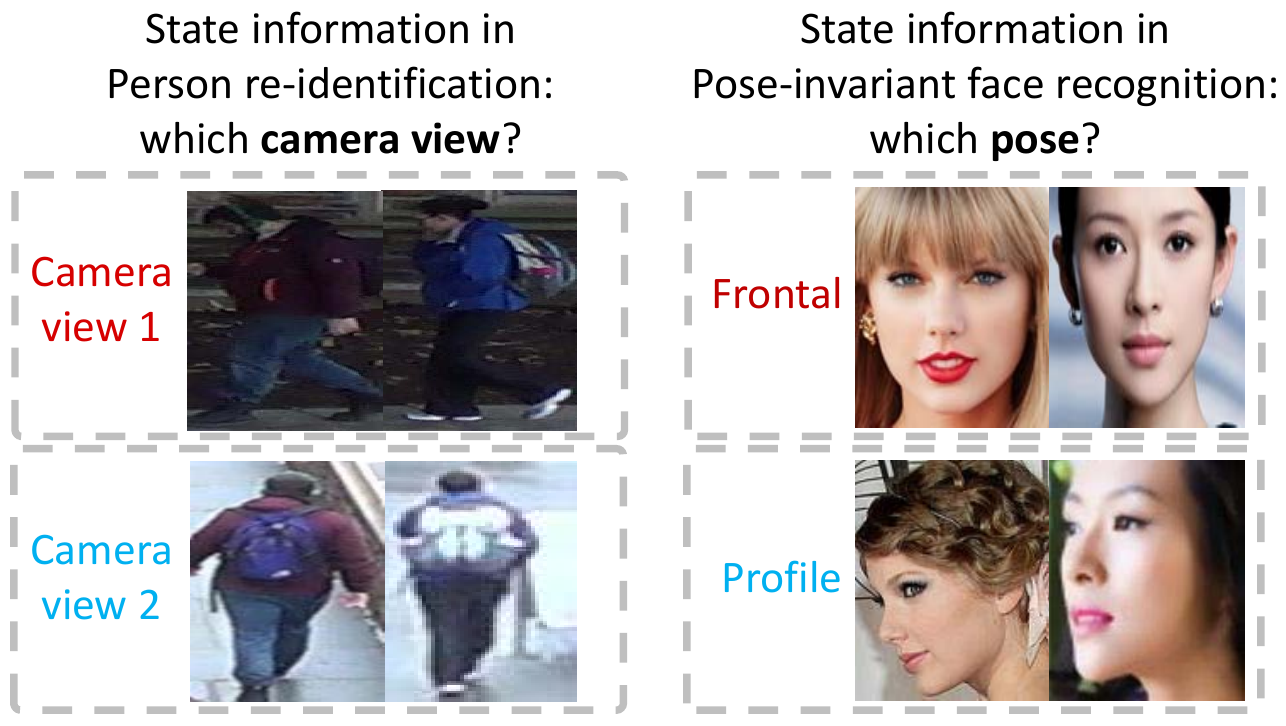}
\end{center}
   \caption{Examples of the state information.
   A pair of images in each column are of the same individual \K{(We do \emph{not} assume to know any pairing;
   this figure is only for demonstrating that different states induce visual discrepancy. We only assume to know the camera/pose label of each image but not the pairing)}.
   }
\label{fig:side_information}
\end{figure}

However, the images of an identity can be drastically different when they are taken under different
states such as different poses and camera views.
For example, we observe great visual discrepancy in the images of the same pedestrian under different camera views in a surveillance scenario (See Figure \ref{fig:side_information}).
Such visual discrepancy caused by the different states
induces great difficulty in unsupervised discriminative learning.
Fortunately, in real-world discriminative tasks,
\K{we can often have some \emph{state information}
without human annotation effort.}
For instance, in person re-identification,
it is straightforward to know from which camera view an unlabeled image is taken \cite{2017_ICCV_asymmetric,2019_CVPR_MAR,2019_TPAMI_DECAMEL,2018_TPAMI_CRAFT},
and in face recognition the pose and facial expression can be estimated by off-the-shelf estimators \cite{MTCNN,shan2009facial} (See Figure \ref{fig:side_information}).
We aim to exploit the state information as weak supervision to address the visual discrepancy in unsupervised discriminative learning.
We refer to our task as the weakly supervised discriminative feature learning.

In this work, we propose a novel pseudo label model for weakly supervised discriminative feature learning.
\K{We assign every unlabeled image example to a surrogate class (i.e. artificially created pseudo class) which is expected to represent an unknown identity in the unlabelled training set,
and we construct the surrogate classification as a simple basic model.
However the unsupervised assignment is often incorrect, because the image features of the same identity are distorted due to the aforementioned visual discrepancy.}
When the visual discrepancy is moderate,
in the feature space,
an unlabeled example ``slips away'' from the correct decision region and
crosses the decision boundary to the decision region of a nearby surrogate class
\kk{(See the middle part in Figure \ref{fig:overview})}.
We refer to this effect as the feature distortion.
We develop the \emph{weakly supervised decision boundary rectification} to address this problem.
The idea is to rectify the decision boundary to encourage the unlabeled example back to the correct decision region.

When the feature distortion is significant, however,
the unlabeled example can be pushed far away from the correct decision region.
Fortunately, the feature distortion caused by a state often follows a specific distortion pattern (e.g., extremely dark illumination in Figure \ref{fig:side_information}
may suppress most visual features).
Collectively, this causes a specific global feature drift (See the right part in Figure \ref{fig:overview}).
Therefore, we alleviate the significant feature distortion to a moderate level (so that it can be addressed by the decision boundary rectification) by countering the global-scale feature drifting.
Specifically, we achieve this by introducing the \emph{weakly supervised feature drift regularization}.

We evaluate our model on two tasks, i.e. unsupervised person re-identification and pose-invariant face recognition.
We find that our model could perform comparably with the standard supervised learning on DukeMTMC-reID \cite{Duke1}, Multi-PIE \cite{MPIE} and CFP \cite{CFP} datasets.
We also find our model could outperform the state-of-the-art unsupervised models on DukeMTMC-reID
and supervised models on Multi-PIE and CFP.
To our best knowledge, this is the first work to develop a weakly supervised discriminative learning model
that can successfully apply to different tasks, leveraging different kinds of state information.

\vspace{-0.2cm}
\section{Related Work}

\noindent
\textbf{Learning with state information}.
State information has been explored separately in identification tasks.
In person re-identification (RE-ID),
several works leveraged the camera view label to help learn view-invariant features and distance metrics
\cite{2015_CVPR_LOMO,2018_TPAMI_CRAFT,2018_TIP_crossview,2018_TPAMI_crossview,2018_CVPR_camera-style}.
In face recognition,
the pose label was also used to learn pose-invariant models
\cite{2018_CVPR_towards,2013_ICCV_random,2014_CVPR_SPAE,2014_NIPS_MVP,2015_CVPR_CPI,2017_CVPR_DRGAN,2018_TIP_multitask}.
\kkk{Specifically, \cite{2019_TPAMI_DECAMEL} and \cite{Peng_2017_ICCV} visualized the feature embedding
to illustrate the feature distortion problem nicely for person re-identification and face recognition,
respectively.}
However, most existing methods \ws{are} based on supervised learning, and thus the prohibitive labelling cost
\ws{could largely} limit their scalability.
Therefore, \emph{unsupervised RE-ID}
\cite{2019_TPAMI_DECAMEL,2017_ICCV_asymmetric,2018_ECCV_tracklet,fan2017unsupervised,2019_CVPR_distill,2019_CVPR_patch,Wu_2019_ICCV}
and \emph{cross-domain transfer learning RE-ID} \cite{2019_CVPR_MAR,2018_CVPR_SPGAN,2018_ECCV_HHL,2019_CVPR_invariance,2018_CVPR_transferable,zhong2019learning,2019_ICCV_CRGAN,2019_ICCV_self-training,Fu_2019_ICCV} have been attracting increasing attention.
These methods typically \ws{incorporate} the camera view labels
to learn the camera view-specific feature transforms \cite{2017_ICCV_asymmetric,2019_TPAMI_DECAMEL},
to learn the soft multilabels \cite{2019_CVPR_MAR},
to provide associations between the video RE-ID tracklets \cite{2018_ECCV_tracklet},
or to generate augmentation images \cite{2018_ECCV_HHL,2019_CVPR_invariance,zhong2019learning}.
\kkk{Our work is different from the cross-domain transfer learning RE-ID methods
in that we do not need any labeled data in the training stage.
As for the unsupervised RE-ID methods,} the most related works are \cite{2017_ICCV_asymmetric,2019_TPAMI_DECAMEL} where Yu et.al. proposed
the asymmetric clustering
in which the camera view labels were leveraged to learn a set of view-specific projections.
However, they need to learn as many projections as the camera views via
solving the costly eigen problem, which limits their scalability. In contrast we learn a generalizable feature for all kinds of states (camera views).

\vspace{0.1cm}
\noindent
\textbf{Weakly supervised learning}.
Our method is to iteratively refine pseudo labels with the state information
which is regarded as weak supervision.
The state information serves to guide the pseudo label assignments as well as to improve the feature invariance against distractive states.

In literatures, weak supervision is a broadly used term. Typical weak supervision \cite{zhou2017weakly} includes image-level coarse labels for finer tasks like detection \cite{bilen2015weakly,bilen2016weakly} and segmentation \cite{wei2016stc,pathak2015constrained}.
Another line of research that is more related to our work is utilizing
large-scale inaccurate labels (typically collected online \cite{2016_CVPR_weakly_supervised_place_recognition} or from a database like Instagram \cite{2018_ECCV_weakly} or Flickr \cite{joulin2016learning}) to learn general features.
Different from existing works, our objective is to learn identity-discriminative features that are directly applicable to identification tasks without supervised fine-tuning.

\vspace{0.1cm}
\noindent
\textbf{Unsupervised deep learning}.
Beyond certain vision applications,
general unsupervised deep learning is a long-standing problem in vision community.
The typical lines of research include
clustering based methods
\cite{2018_ECCV_deep-clustering,2016_CVPR_deep-clustering,2016_NIPS_cliquecnn,2014_NIPS_surrogate}
which discovered cluster structures in the unlabelled data and utilized the cluster labels,
and the generation based methods
which learned low-dimensional features that were effective for generative discrimination \cite{DCGAN,donahue2016adversarial,2014_NIPS_GAN} or reconstruction \cite{denoising_autoencoder,vae,bengio2007greedy}.

Recently, self-supervised learning, a promising paradigm of unsupervised learning,
has been quite popular.
Self-supervised methods typically construct some pretext tasks
where the supervision comes from the data.
Typical pretext tasks include predicting relative patch positions \cite{2015_ICCV_context-prediction},
predicting future patches \cite{oord2018representation},
solving jigsaw puzzles \cite{2016_ECCV_unsupervised,jigsawplus},
image inpainting \cite{2016_CVPR_inpainting},
image colorization \cite{2016_ECCV_colorization,zhang2017split}
and predicting image rotation \cite{2018_ICLR_predicting_rotation}.
%Besides images, self-supervision from videos included synchronized multi-modality data
%of audio, video frames and language text \cite{wang2019reinforced,gomez2017self,wiles2018self,owens2018audio,korbar2018cooperative},
%and temporal consistency from videos \cite{sermanet2018time}.
By solving the pretext tasks,
they aimed to learn features that were useful for downstream real-world tasks.

Our goal is different from these works.
Since they \ws{aim} to learn useful features for various downstream tasks,
they were designed to be downstream task-agnostic,
and required supervised fine-tuning for them.
In contrast, we actually focus on the ``fine-tuning'' step, with a goal to reduce the need of manual labeling.

%
%\noindent
%\textbf{Multi-view learning}.
%\ws{Our model is different from multi-view learning, although they seem related. Multi-view learning} considers learning from multiple modalities/views of the same example \cite{multiview_survey,multiview_survey2}.
%Images of different states could be seen as different views.
%\K{However, the multi-view learning models assume that the \emph{pairing} of different views is known, 
%which is supervised learning in our context.
%In contrast, in our unsupervised setting we only know the state of each example but not the pairing of the unlabeled data.
%}

\section{\ws{Weakly supervised Discriminative Learning with State Information}}

\begin{figure*}[t]
\begin{center}
\includegraphics[width=0.85\linewidth]{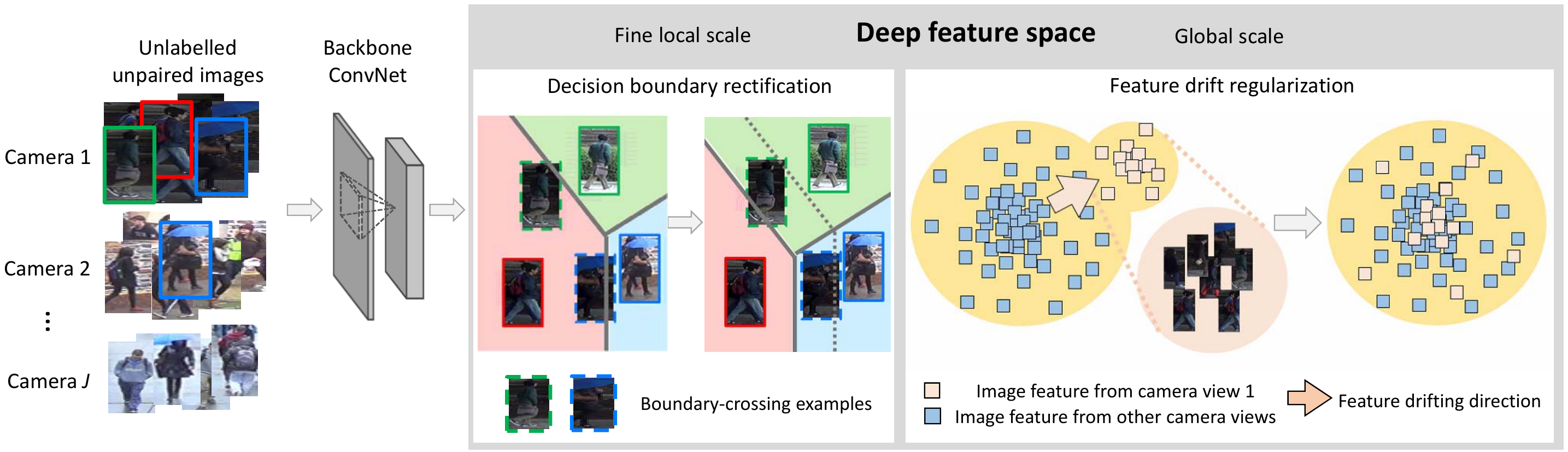}
\vspace{-0.5cm}
\end{center}
   \caption{An illustration in which the weak supervision is the \emph{camera view} in person re-identification.
   In the decision boundary rectification, each colored bounding box denotes an identity
   (the identity labels are \emph{unknown}; we use them here only for illustrative purpose).
   Each colored region denotes the decision region of a surrogate class. 
   \kk{In the feature drift regularization, we illustrate the visually dominant state by the dark camera view 1.}
   Best viewed in color.
   }
\label{fig:overview}
\vspace{-0.5cm}
\end{figure*}

Let $\mathcal{U}=\{u_i\}_{i=1}^{N}$ denote the unlabelled training set,
where $u_i$ is an unlabelled image example.
We also know the state $s_i \in \{1,\cdots, J\}$,
e.g., the illumination of $u_i$ is dark, normal or bright.
Our goal is to learn a deep network $f$ to extract identity-discriminative feature
which is denoted by $x = f(u; \theta)$.
A straightforward idea is to assume that in the feature space every $x$
belongs to a surrogate class which is modelled by a surrogate classifier $\mu$.
A surrogate class is expected to model a potential unknown identity in the unlabeled training set.
The discriminative learning can be done by a surrogate classification:
{\small
\begin{align}\label{eq:loss_aa}
\min_{\theta, \{\mu_k\}} L_{surr} = -\Sigma_{x} \log\frac{\exp(x^\mathrm{T}\mu_{\hat{y}})}{\Sigma_{k=1}^K \exp(x^\mathrm{T}\mu_k)},
\end{align}
}%
where $\hat{y}$ denotes the surrogate class label of $x$, and $K$ denotes the number of surrogate classes.
An intuitive method for surrogate class assignment is:
{\small
\begin{align}\label{eq:assignment}
\hat{y} = \arg\max_k \: \exp(x^\mathrm{T}\mu_{k}).
\end{align}
}%
% Apparently, the underlying assumption is that the visual appearance features of the same entity
% should be similar.
However, the visual discrepancy caused by the state leads to incorrect assignments.
When the feature distortion is moderate,
wrong assignments happen locally, i.e., $x$ wrongly crosses the decision boundary into a nearby surrogate class' decision region.
We develop the \emph{Weakly supervised Decision Boundary Rectification} (WDBR) to address it.
As for the \emph{significant} feature distortion, however,
it is extremely challenging as $x$ is pushed far away from the correct decision region.
To deal with it, we introduce the \emph{Weakly supervised Feature Drift Regularization} to
alleviate the significant feature distortion down to a moderate level that WDBR can address.
We show an overview illustration in Figure \ref{fig:overview}.

\subsection{Weakly supervised decision boundary rectification (WDBR)}

\begin{figure}[t]
\begin{center}
\subfigure[Rectifier function]{
\includegraphics[width=0.45\linewidth,height=0.25\linewidth]{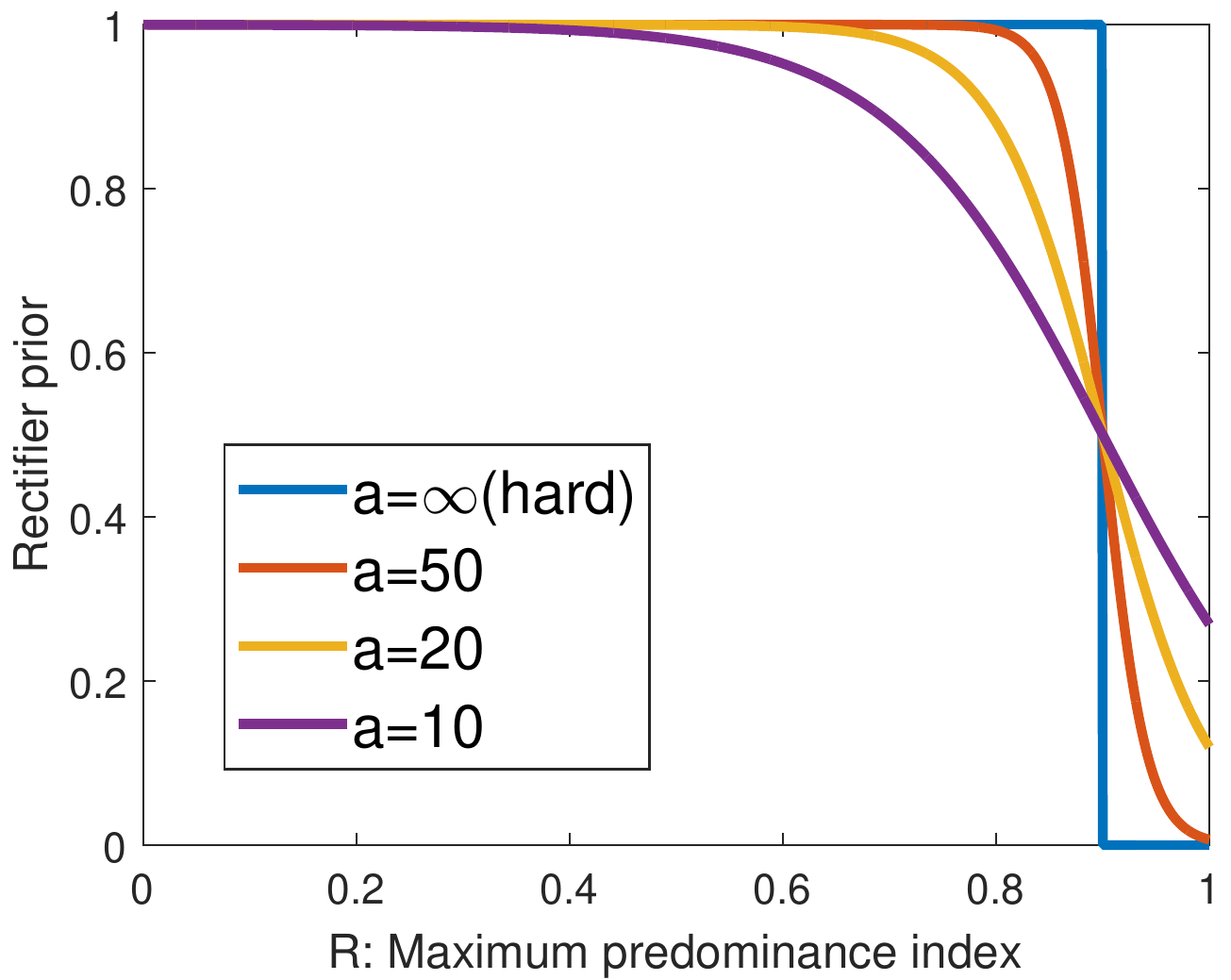}\label{fig:prior}
}
\subfigure[A set of decision boundaries]{
\includegraphics[width=0.45\linewidth,height=0.25\linewidth]{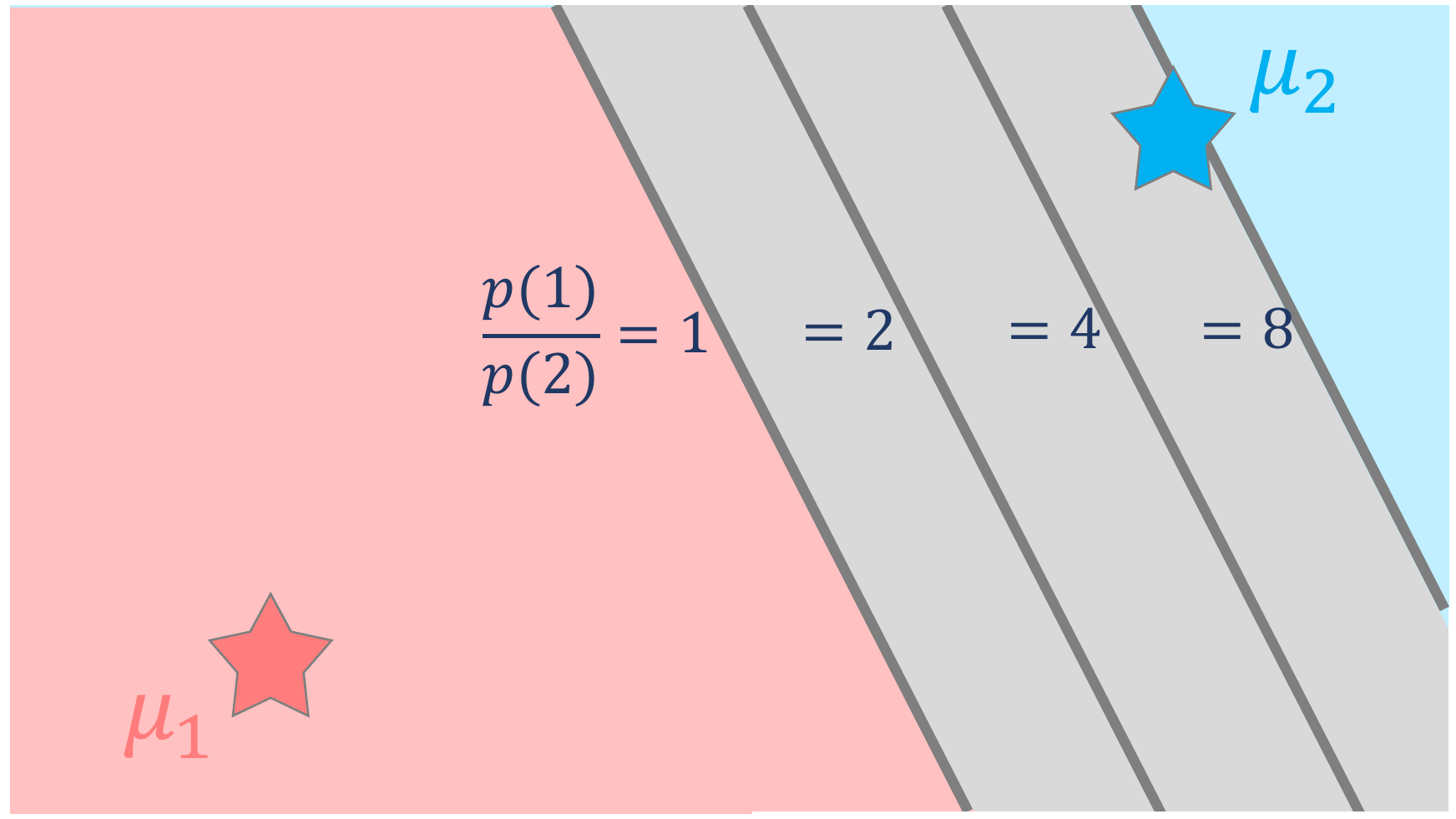}\label{fig:decision_boundary}
}
\end{center}
\caption{(a) The rectifier function $p(k)$ with different strength $a$.
(b) Decision boundaries with different $\frac{p(1)}{p(2)}$ values.}
\vspace{-0.4cm}
\end{figure}

We first consider the moderate visual feature distortion.
It ``nudges'' an image feature $x$ to wrongly cross the decision boundary
into a nearby surrogate class.
For example, two persons wearing dark clothes are even harder to distinguish
when they both appear in a dark camera view.
Thus, these person images are assigned to the same surrogate class (see Figure \ref{fig:overview} for illustration).
In this case, a direct observation
is that most members of the surrogate class is taken from the same dark camera view (i.e. the same state).
Therefore, we quantify the extent to which a surrogate class is dominated by a state.
We push the decision boundary toward a highly dominated surrogate class \kk{or even nullify it, }
in an attempt to correct these local boundary-crossing wrong assignments.

We quantify the extent by the \emph{Maximum Predominance Index} (MPI).
The MPI is defined as the proportion of the most common state in a surrogate class.
Formally, the MPI of the $k$-th surrogate class $R_k$ is defined by:
{\small
\begin{align}\label{eq:AFPI}
R_k = \frac{\max_j|\mathcal{M}_k\cap \mathcal{Q}_j|}{|\mathcal{M}_k|} \in [0,1],
\end{align}
}%
where the denominator is the number of members in a surrogate class,
formulated by the cardinality of the \emph{member set} of the $k$-th surrogate class $\mathcal{M}_k$:
{\small
\begin{align}\label{eq:member_set}
\mathcal{M}_k = \{x_i| \hat{y}_i = k\},
\end{align}
}%
and the numerator is the number of presences of the most common state in $\mathcal{M}_k$.
We formulate it by the intersection of $\mathcal{M}_k$ and the \emph{state subset} 
corresponding to the $j$-th state $\mathcal{Q}_j$:
{\small
\begin{align}\label{eq:partial_set}
\mathcal{Q}_j = \{x_i| s_i = j\}.
\end{align}
}%
Note that the member set $\mathcal{M}_k$ is dynamically updated,
as the surrogate class assignment (Eq. (\ref{eq:assignment})) is on-the-fly along with the learning,
and is improved upon better learned features.
%\kk{We define $R_k =0$ when no member is assigned to it.}

As analyzed above, a higher $R_k$ indicates that it is more likely that some examples
have wrongly crossed the decision boundary into the surrogate class $\mu_k$ due to the feature distortion.
Hence, we shrink that surrogate class' decision boundary to purge the potential boundary-crossing examples from its decision region.
Specifically,
we develop the weakly supervised rectified assignment:
{\small
\begin{align}\label{eq:assignment_2}
\hat{y} = \arg\max_k \: p(k)\exp(x^\mathrm{T}\mu_{k}),
\end{align}
}%
where $p(k)$ is the rectifier function that is monotonically decreasing with $R_k$:
{\small
\begin{align}\label{eq:prior}
p(k) = \frac{1}{1+\exp(a\cdot(R_k-b))} \in [0,1],
\end{align}
}%
where $a\geq 0$ is the rectification strength and $b\in[0,1]$ is the rectification threshold.
\kk{We typically set $b=0.95$.
In particular, we consider $a=\infty$, and thus we have:
{\small
\begin{align}\label{eq:prior_2}
p(k) =
\begin{cases}
1, & \text{if } R_k \leq b \\
0, &\text{otherwise}
\end{cases}
\end{align}
}%
This means that when the MPI exceeds the threshold $b$ we nullify it by shrinking its decision boundary to a single point.}
We show a plot of $p(k)$ in Figure \ref{fig:prior}.

For any two neighboring surrogate classes $\mu_1$ and $\mu_2$,
the decision boundary is (\ws{where} we leave the derivation to the supplementary material):
{\small
\begin{align}
(\mu_1-\mu_2)^\mathrm{T}x + \log\frac{p(1)}{p(2)} = 0.
\end{align}
}%
%which is a line perpendicular to $\mu_1-\mu_2$, moved according to $\log\frac{p(2)}{p(1)}$.
% \vspace{0.1cm}
\kk{
\noindent
\textbf{Discussion}.
To have a better understanding of the WDBR, let us first consider the hard rectifier function.
When a surrogate class' MPI exceeds the threshold $b$ (typically we set $b=0.95$), the decision region vanishes,
and no example would be assigned to the surrogate class (i.e., it is completely nullified).
Therefore, WDBR prevents the unsupervised learning
from being misled by those severely affected surrogate classes.
For example, if over $95\%$ person images assigned to a surrogate class
are from the same dark camera view,
it is highly likely this is simply because
it is too dark to distinguish them, rather than because they are the same person.
Thus, WDBR nullifies this poorly formed surrogate class.

When we use the soft rectifier function, the WDBR does not directly nullify the surrogate class that exceeds the threshold, but favors the surrogate class which has lower MPI (because they are less likely to have the boundary-cross problem) by moving the decision boundary.
This can be seen from Figure \ref{fig:decision_boundary} where we plot a set of decision boundaries in the two-class case.
In some sense, the soft WDBR favors the state-balanced surrogate classes.
This property may further improve the unsupervised learning,
especially if the unlabelled training set is indeed state-balanced for most identities.
However, if we do not have such a prior knowledge of state balance, using hard rectifier can be more desirable, because hard rectifier does not favor state-balanced surrogate classes.
We will discuss more about this property upon real cases in Sec. \ref{sec:ablation}.
}

In the supplementary material, we theoretically justify our model by showing that
the rectified assignment is the maximum a posteriori optimal estimation of $\hat{y}$.
However, the WDBR is a local mechanism, i.e. WDBR deals with the moderate feature distortion that
nudges examples to slip in nearby surrogate classes.
Its effectiveness might be limited when the feature distortion is significant.

\subsection{Weakly supervised feature drift regularization}

A visually dominant state may cause a \emph{significant} feature distortion
that pushes an example far away from the correct surrogate class.
This problem is extremely difficult to address by only considering a few surrogate classes in a local neighborhood.
Nevertheless, such a significant feature distortion is likely to follow a specific pattern.
For example, the extremely low illumination may suppress all kinds of visual features:
dim colors, indistinguishable textures, etc.
Collectively, we can capture the significant feature distortion pattern in a global scale.
In other words, such a state-specific feature distortion
would cause many exmaples $x$ in the state subset to drift toward a specific direction (see Figure \ref{fig:overview} for illustration).
We capture this by the state sub-distribution and introduce the Weakly supervised Feature Drift Regularization (WFDR) to address it
and complement the WDBR.

In particular, we define the \emph{state sub-distribution} as $\mathbb{P}(\mathcal{Q}_j)$, 
which is the distribution over the state subset $\mathcal{Q}_j$ defined in Eq. (\ref{eq:partial_set}).
For example, all the unlabeled person images captured from a dark camera view.
We further denote the distribution over the whole unlabelled training set as $\mathbb{P}(\mathcal{X})$,
where $\mathcal{X}=f(\mathcal{U})$.
Apparently, the state-specific feature distortion would lead to a specific
sub-distributional drift, i.e., $\mathbb{P}(\mathcal{Q}_j)$ drifts away from $\mathbb{P}(\mathcal{X})$.
For example, all person images from a dark camera view may be extremely low-valued in many feature dimensions,
and this forms a specific distributional characteristic.
Our idea is straightforward:
we counter this ``collective drifting force'' by aligning the state sub-distribution $\mathbb{P}(\mathcal{Q}_j)$ with the overall total distribution $\mathbb{P}(\mathcal{X})$ to suppress the significant feature distortion.
We formulate this idea as the Weakly supervised Feature Drift Regularization (WFDR):
{\small
\begin{align}\label{eq:distribution}
\min_{\theta} L_{drift} = \Sigma_{j} d(\mathbb{P}(\mathcal{Q}_j), \mathbb{P}(\mathcal{X})),
\end{align}
}%
where $d(\cdot, \cdot)$ is a distributional distance.
In our implementation
we \ws{adopt} the simplified 2-Wasserstein distance \cite{2017_Arxiv_BEGAN,2018_TPAMI_WCNN} as $d(\cdot,\cdot)$
due to its simplicity and computational ease.
In particular, it is given by:
{\small
\begin{align}
d(\mathbb{P}(\mathcal{Q}_j), \mathbb{P}(\mathcal{X})) = ||m_j - m||_2^2 + ||\sigma_j - \sigma||_2^2,
\end{align}
}%
where $m_j$/$\sigma_j$ is the mean/standard deviation feature vector over $\mathcal{Q}_j$.
Similarly, $m$/$\sigma$ is the mean/standard deviation feature vector over the whole unlabelled training set $\mathcal{X}$.

Ideally, WFDR alleviates the significant feature distortion down to a mild level (i.e., $x$ is regularized into the correct decision region)
or a moderate level (i.e., $x$ is regularized into the neighborhood of the correct surrogate class) that the WDBR can address.
Thus, it is mutually complementary to the WDBR.
We note that the WFDR is mathematically akin to the soft multilabel learning loss
in \cite{2019_CVPR_MAR},
but they serve for different purposes.
The soft multilabel learning loss is to align
the cross-view \emph{associations} between unlabeled target images and labeled source images,
while we aim to align the feature distributions of unlabeled images and we do not need a source dataset.

%In our implementation, we maintain a buffer for $m$ and $\sigma$ as a reference,
%whereas $m_j$ and $\sigma_j$ are estimated within each batch to obtain the gradient.
%Similar to the batch normalization \cite{2015_Arxiv_BN}, we updated the buffer with a momentum $\alpha=B/N$ for each batch,
%where $B$ is the batch size and $N$ is the training set size (we shall show details later in Algorithm \ref{alg:sasi}).
Finally, the loss function of our model is:
{\small
\begin{align}\label{eq:loss_total}
\min_{\theta, \{\mu_k\}} L = L_{surr} + \lambda L_{drift},
\end{align}
}%
where $\lambda>0$ is a hyperparameter to balance the two terms.

In our implementation
we used the standard ResNet-50 \cite{2016_CVPR_ResNet} as our backbone network.
We trained our model for approximately 1,600 iterations with batchsize 384,
momentum 0.9 and weight decay 0.005.
We followed \cite{2016_CVPR_ResNet} to use SGD, set the learning rate to 0.001,
and divided the learning rate by 10 after 1,000/1,400 iterations.
We \ws{used} a single SGD optimizer for both $\theta$ and $\{\mu_k\}_{k=1}^K$.
Training costed less than two hours by using 4 Titan X GPUs.
We initialized the surrogate classifiers $\{\mu_k\}_{k=1}^K$
by performing standard K-means clustering on the initial feature space and using the cluster centroids.
For further details please refer to the supplementary.
We also summarize our method in an algorithm in the supplementary material.

\section{Experiments}
\subsection{Datasets}
\begin{figure}[t]
\begin{center}
\subfigure[Market-1501]{
\includegraphics[width=0.4\linewidth,height=0.15\linewidth]{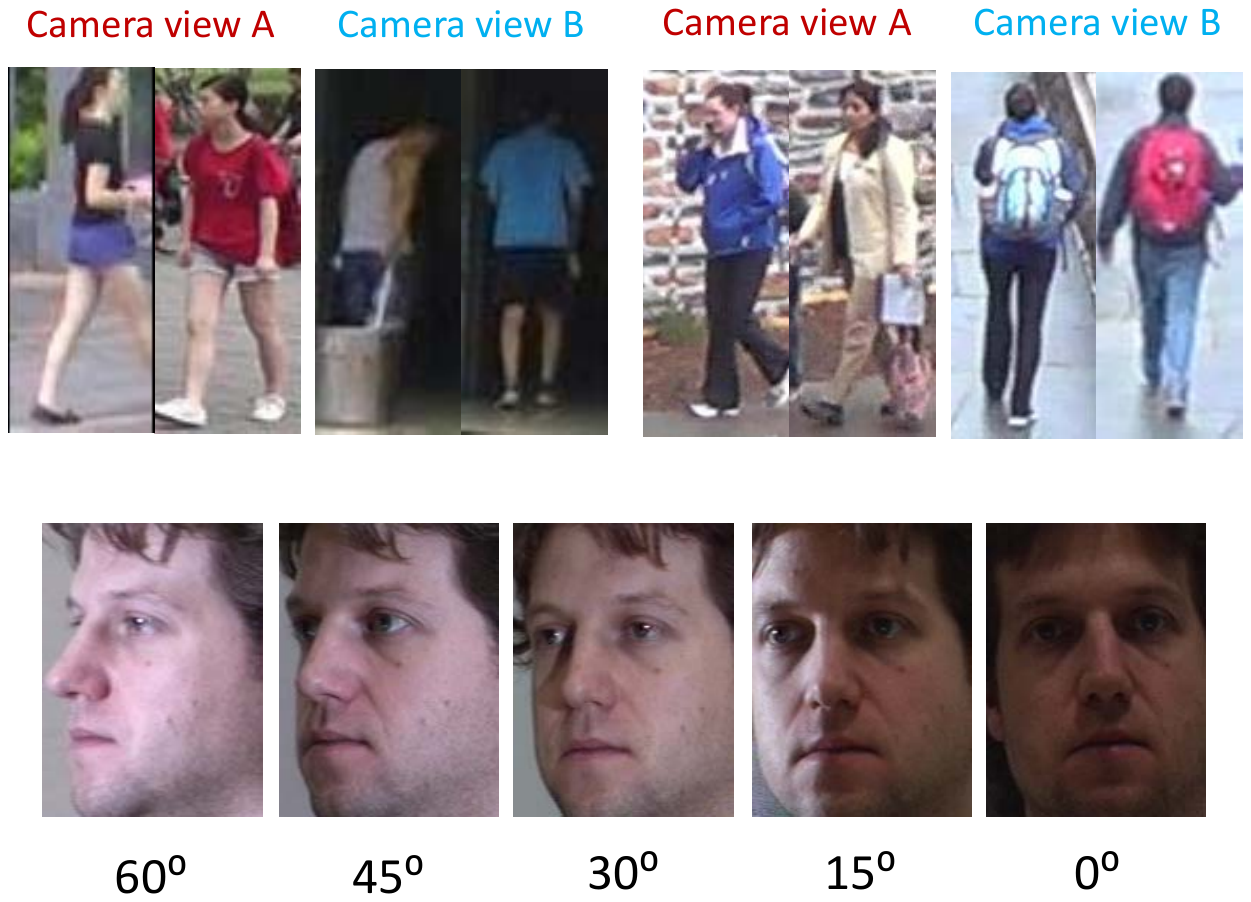}\label{fig:dataset_market}
}
\vspace{-0.3cm}
\subfigure[DukeMTMC-reID]{
\includegraphics[width=0.4\linewidth,height=0.15\linewidth]{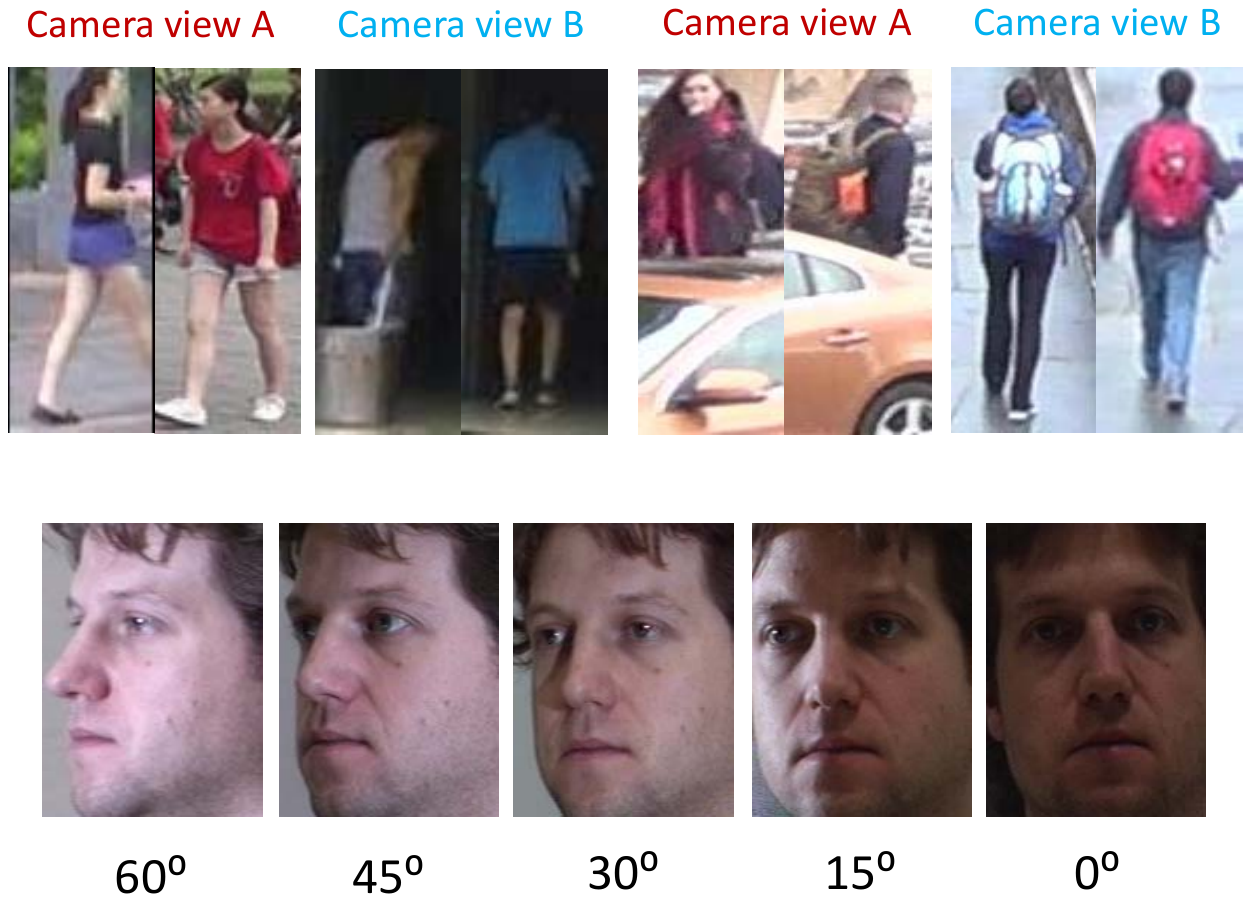}\label{fig:dataset_duke}
}
\vspace{-0.3cm}
\subfigure[Multi-PIE]{
\includegraphics[width=0.55\linewidth,height=0.15\linewidth]{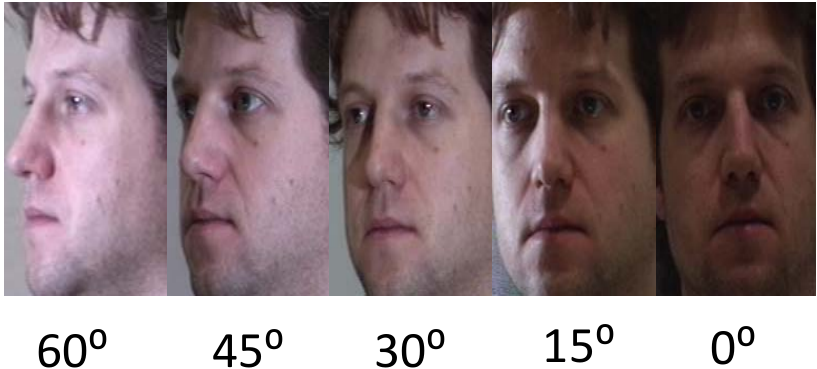}\label{fig:dataset_mpie}
}
\vspace{-0.1cm}
\subfigure[CFP]{
\includegraphics[width=0.25\linewidth,height=0.15\linewidth]{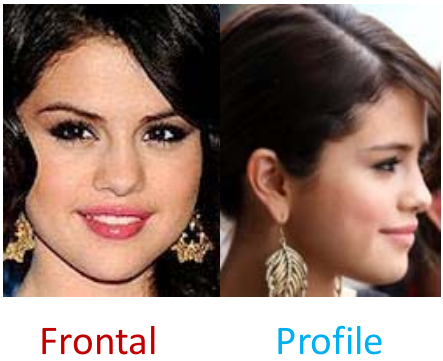}\label{fig:dataset_cfp}
}
\vspace{-0.1cm}
\end{center}
   \caption{Dataset examples. The state information for RE-ID and PIFR is camera view labels and pose labels, respectively.
   }
\label{fig:dataset_examples}
\vspace{-0.2cm}
\end{figure}

We evaluated our model on two real-world discriminative tasks with state information,
i.e. person re-identification (RE-ID) \cite{2016_Arxiv_Zheng} and pose-invariant face recognition (PIFR) \cite{2000_FG_pose-invariant,2016_TIST_pose-invariant}.
In RE-ID which aims to match person images across non-overlapping camera views,
the state information is the camera view label, as illustrated in Figure \ref{fig:dataset_market} and \ref{fig:dataset_duke}.
Note that each camera view has its specific conditions including illumination, viewpoint and occlusion (e.g. Figure \ref{fig:dataset_market} and \ref{fig:dataset_duke}).
In PIFR, which aims to identify faces across different poses,
the state information is the pose, as illustrated in Figure \ref{fig:dataset_mpie}.
We note that on both tasks the training identities are
completely different from the testing identities.
Hence, these tasks are suitable to evaluate the discriminability and generalisability of learned feature.

\vspace{0.1cm}
\noindent
\textbf{Person re-identification (RE-ID)}.
We evaluated on Market-1501 \cite{2015_ICCV_MARKET} and DukeMTMC-reID \cite{Duke1,Duke2}.
Market-1501 contains 32,668 person images of 1,501 identities.
Each person is taken images from at least 2 out of 6 disjoint camera views.
We followed the standard evaluation protocol where the training set had 750 identities and testing set had the other 751 identities \cite{2015_ICCV_MARKET}.
The performance was measured by the cumulative accuracy and the mean average precision (MAP) \cite{2015_ICCV_MARKET}.
DukeMTMC-reID contains 36,411 person images of 1,404 identities.
Images of each person were taken from at least 2 out of 8 disjoint camera views.
We followed the standard protocol which was similar to the Market-1501 \cite{Duke2}.
We followed \cite{2019_CVPR_MAR} to pretrain the network with standard softmax loss on the MSMT17 dataset \cite{2018_CVPR_PTGAN}
in which the scenario and identity pool were completely different from Market-1501 and DukeMTMC-reID.
It should be pointed out that in fine-grained discriminative tasks like RE-ID and PIFR,
the pretraining is important for unsupervised models because the class-discriminative visual clues
are not general but highly task-dependent \cite{2016_Arxiv_transferREID,2017_Arxiv_PUL,2018_CVPR_transferable,2019_TPAMI_DECAMEL},
and therefore some extent of field-specific knowledge is necessary for successful unsupervised learning.
We resized the images to $384 \times 128$.
In the unsupervised setting, the precise number of training classes (persons) $P$ (i.e. 750/700 for Market-1501/DukeMTMC-reID) should be unknown.
Since our method was able to automatically discard excessive surrogate classes,
an ``upper bound'' estimation could be reasonable.
We set $K=2000$ for both datasets.

\vspace{0.1cm}
\noindent
\textbf{Pose-invariant face recognition (PIFR)}.
We mainly evaluated on the large dataset Multi-PIE \cite{MPIE}.
Multi-PIE contains 754,200 images of 337 subjects taken with
up to 20 illuminations, 6 expressions and 15 poses \cite{MPIE}.
For Multi-PIE, most experiments followed the widely-used setting \cite{2014_NIPS_MVP} which used all 337 subjects with neutral expression
and 9 poses interpolated between $-60\degree$ and $60\degree$.
The training set contained the first 200 persons,
and the testing set contained the remaining 137 persons.
When testing, one image per identity with the frontal view was put into the gallery set
and all the other images into the query set.
The performance was measured by the top-1 recognition rate.
We detected and cropped the face images by MTCNN \cite{MTCNN},
resized the cropped images to $224\times 224$,
and we adopted the pretrained model weights provided by \cite{VGGface2}.
Similarly to the unsupervised RE-ID setting, we simply set $K=500$.
We also evaluated on an unconstrained dataset CFP \cite{CFP}.
The in-the-wild CFP dataset contains 500 subjects with 10 frontal and 4 profile images for each subject.
We adopted the more challenging frontal-profile verification setting \cite{CFP}.
We followed the official protocol \cite{CFP}.
to report the mean accuracy, equal error rate (EER) and area under curve (AUC).

In the unsupervised RE-ID task, the camera view labels were naturally available \cite{2019_CVPR_MAR,2019_TPAMI_DECAMEL}.
In PIFR we used groundtruth pose labels for better analysis.
In the supplementary material we showed the simulation results when we used the estimated pose labels.
The performance did not drop until the correctly estimated pose labels were less than $60\%$.
\Koven{In practice the facial pose is continuous and we need to discretize it to produce the pose labels.
In our preliminary experiments on Multi-PIE we found that merging the pose labels into coarse-grained
groups did not affect the performance significantly.
Therefore, for fair comparison to other methods, we followed the conventional setting to use the default pose labels.}
We set $\lambda = 10$ and $b=0.95$ for all datasets except Multi-PIE which has
more continual poses and thus we decreased to $\lambda = 1$, $b=0.5$.
We evaluated both soft version $a=5$ and hard version $a=\infty$.
We provide evaluations and analysis for $K$,$\lambda$,$a$ and $b$ in the supplementary material.

\begin{table}[t]
\centering
\scriptsize
\caption{\label{tb:ablation_reid}
Model evaluation on the person re-identification (\%).
Please refer to Sec. \ref{sec:ablation} for description of the compared methods.
}
\begin{tabular}{c|ccc|ccc}
\hline
\multirow{2}*{Methods}  & \multicolumn{3}{c}{DukeMTMC-reID} & \multicolumn{3}{c}{Market-1501} \\
\cline{2-7}
& rank-1 & rank-5 & mAP & rank-1 & rank-5 & mAP \\
\hline
Supervised fine-tune &75.0&85.0&57.2&85.9&95.2&66.8\\
\hline
Pretrained & 43.1 & 59.2 & 28.8&46.2 & 64.4 & 24.6 \\
K-means as labels  &37.3&52.1&25.2&47.3&63.1& 25.6 \\
Basic model &56.8&71.1&39.6&55.8 &72.2&31.5  \\
Basic + WDBR (hard) &69.4&80.5&50.2&60.3&73.4&34.5 \\
Basic + WDBR (soft) &63.6&77.2&45.4&60.0&75.6&34.3\\
Basic + WFDR &67.7&79.4&47.5&67.4&82.3&39.4 \\
\hline
Full model (hard) &\textbf{72.1} &\textbf{83.5}&\textbf{53.8}&\textbf{74.0}&\textbf{87.4} &\textbf{47.9}\\
Full model (soft) &70.3&81.7&50.0&70.7&85.2&43.4\\
\hline
\end{tabular}
\vspace{-0.2cm}
\end{table}

\begin{table}[t]
\centering
\scriptsize
\caption{\label{tb:ablation_PIFR}
Model evaluation on the Multi-PIE (\%).
We report the mean results averaged over 5 runs.
Please refer to Sec. \ref{sec:ablation} for description of the compared methods.
}
\begin{tabular}{c|c|ccccc}
\hline
Methods & avg & 0\degree & $\pm$15\degree & $\pm$30\degree & $\pm$45\degree & $\pm$60\degree \\
\hline
Supervise fine-tune &98.2&99.7&99.4&98.8&98.1&95.7\\
\hline
Pretrained &88.7&98.5&97.5&93.7&89.7&71.2 \\
K-means as labels &81.0 &95.7&94.6&89.1&76.7&56.0\\
Basic model &54.5&91.2&86.5&60.0&34.5&18.8\\
Basic + WDBR (hard)  &91.7&98.9&98.7&97.5&91.2&75.9\\
Basic + WDBR (soft) &97.0&99.1&98.9&98.3&96.8&93.0 \\
Basic + WFDR  &95.7&98.4&98.1&97.0&95.5&91.0\\
\hline
Full model (hard) & 95.7&98.3&98.1&97.0&95.3&91.1\\
Full model (soft) &\textbf{97.1}&\textbf{99.1}&\textbf{98.9}&\textbf{98.3}&\textbf{96.8}&\textbf{93.1}\\
\hline
\end{tabular}
\vspace{-0.2cm}
\end{table}

\begin{table}[t]
\centering
\scriptsize
\caption{\label{tb:ablation_CFP}
Model evaluation on CFP (\%).
Please refer to Sec. \ref{sec:ablation} for description of the compared methods.
}
\begin{tabular}{c|ccc}
\hline
Methods & Accuracy & EER & AUC \\
\hline
Supervised fine-tune &95.50(0.98)&4.74(1.05)&98.82(0.50)\\
\hline
Pretrained & 92.90(1.37)&7.40(1.37)&97.75(0.73) \\
Basic model &93.57(1.32)&6.89(1.51)&97.55(0.93) \\
\hline
Full model (soft) &\textbf{95.49}(0.70)&\textbf{4.74}(0.72)&\textbf{98.83}(0.29) \\
\hline
\end{tabular}
\vspace{-0.2cm}
\end{table}

\subsection{Model evaluation and analysis}\label{sec:ablation}

We decomposed our model for analysis.
To ground the performance,
we provided the standard supervised fine-tuning results (i.e. replacing our proposed loss with softmax loss with groundtruth class labels,
and keeping other settings the same)
which could be seen as an upper bound.
As an unsupervised baseline, we used K-means cluster labels (i.e. we performed K-means once on the pretrained feature space to obtain the cluster labels,
and used the cluster labels instead of the groundtruth labels for fine-tuning) and we denote this as ``K-means as labels''.
We also ablated both WDBR and WFDR from our full model to obtain a ``Basic model''. The key difference between ``K-means as labels'' and ``Basic model'' is that the former uses fixed cluster labels while the latter dynamically infers pseudo labels every batch along with model training.
We show the results in Table \ref{tb:ablation_reid}, \ref{tb:ablation_PIFR} and \ref{tb:ablation_CFP}.
On CFP the observation was similar to Multi-PIE and we show the most significant results only.

\vspace{0.1cm}
\noindent
\textbf{Comparable performance to standard supervised fine-tuning}.
Compared to the standard supervised fine-tuning,
we found that our model could perform \emph{comparably with the supervised results}
in both the person re-identification task on DukeMTMC-reID and the face recognition task on Multi-PIE and CFP.
The overall effectiveness was clear when we ground the performance by
both the supervised results and the pretrained baseline results.
For example, on DukeMTMC-reID, the supervised learning improved the pretrained network by 31.9\% in rank-1 accuracy,
while our model improved it by 29.0\%, leaving only a gap of 2.9\%.
On Multi-PIE our model achieved 97.1\% average recognition rate which was very closed to the supervised result 98.2\%.
On CFP our model even achieved approximately the same performance as supervised fine-tuning,
probably because the small training set (6300 images) favored a regularization.
We also notice that significant performances held both when
the initial pretrained backbone network was \emph{weak} (e.g. in RE-ID the initial rank-1 accuracy performance was below 50\%)
and when the initial backbone network was \emph{strong} (i.e. in PIFR the initial recognition accuracy performance was over 80\%).
These comparisons verified the effectiveness of our model.

\vspace{0.1cm}
\noindent
\textbf{Soft vs. hard decision boundary rectification}.
We found that the soft rectification performed better
on PIFR benchmarks while hard rectification excelled at RE-ID.
We assumed that a key reason was that on the RE-ID datasets,
different persons' images were \emph{unbalanced}, i.e., some IDs appeared only in two camera views
while some may appear in up to six camera views.
For example, for a person $A$ who appeared in 2 camera views,
the MPI $R_A$ was at least 1/2, while this lower bound was 1/6 for another person who appeared in 6 camera views.
Thus the soft rectifier may unfairly favor the surrogate class corresponding to the person appearing in more camera views.
While the hard rectifier does not favor state-balance:
it only nullified highly likely incorrect surrogate classes with very high MPI.
Therefore, the hard rectification could be more robust to the state imbalance.
On the other hand, for Multi-PIE and CFP where the classes were balanced,
soft rectification would fine-tune the decision boundary to a better position,
and thus achieved better results.
\emph{Hence, in this paper we used the hard WDBR for RE-ID and the soft WDBR for PIFR}.

\vspace{0.1cm}
\noindent
\textbf{Complementary nature of WDBR and WFDR}.
Comparing the basic model (our model without WDBR or WFDR)
to basic model with either WDBR
or WFDR,
the performance was consistently improved.
With both WDBR and WFDR, the performance was further improved.
This showed that the fine local-scale WDBR and the global-scale WFDR were complementarily effective.

We noticed that
on Multi-PIE this complementary nature was less significant,
as using WDBR alone could achieve similar results to the full model.
This may be due to the \emph{continual} nature of the pose variation on Multi-PIE:
the variation from 0\degree to 60\degree is in a locally connected \emph{manifold} \cite{2000_science_isomap},
with 15\degree/30\degree/45\degree in between.
Therefore, it was easier for our local mechanism to gradually ``connect''
some 0\degree/15\degree surrogate classes with some 15\degree/30\degree surrogate classes
to finally have a global aligning effect.
In contrast, in RE-ID this manifold nature is less apparent since it lacks evidence of inherent relations between each pair of camera views.

%We also noticed that the sole WDBR brought much less improvements on Market-1501 than DukeMTMC-reID.
%As the network was pretrained on MSMT17 dataset where pedestrians wore thick and long winter clothes, 
%the pretrained feature embedding was invariant to cross-view appearance variation 
%for the persons in thick long clothes across different camera views. 
%Thus, for DukeMTMC where persons also wore thick long clothes,
%the local mechanism could solely work well (similar to the analysis on Multi-PIE).
%While for Market-1501 where the persons wore light and short summer clothes,
%this good property may not hold and thus the WDBR did not work solely well.

\begin{table}[t]
\centering
\scriptsize
\caption{\label{tb:stoa_REID}
Comparison to the state-of-the-art unsupervised RE-ID (upper) and domain adaptation RE-ID (middle) models.
%\red{\textbf{Red}}: the best. \blue{\textbf{Blue}}: the second best.
}
\begin{tabular}{c|c|cc|cc}
\hline
\multirow{2}*{Methods} & \multirow{2}*{Reference} & \multicolumn{2}{c}{DukeMTMC-reID} & \multicolumn{2}{c}{Market-1501}\\
\cline{3-6}
&& rank-1 & mAP &rank-1 & mAP \\
\hline
\hline
CAMEL \cite{2017_ICCV_asymmetric}& ICCV'17& 40.3  & 19.8& 54.5 & 26.3  \\
PUL \cite{2017_Arxiv_PUL} & ToMM'18 & 30.0 & 16.4& 45.5  & 20.5 \\
DECAMEL \cite{2019_TPAMI_DECAMEL} & TPAMI'19 &-&-& 60.2& 32.4 \\
Distill \cite{Wu_2019_CVPR} & CVPR'19 & 48.4 & 29.4 & 61.5 & 33.5 \\
Wu et.al. \cite{2019_ICCV_Wu} & ICCV'19 & 59.3&37.8&65.4&35.5\\

\hline
HHL \cite{2018_ECCV_HHL} & ECCV'18 & 46.9  & 27.2& 62.2 & 31.4 \\
ECN \cite{2019_CVPR_invariance} & CVPR'19  & 63.3 & 40.4&75.1 & 43.0 \\
MAR \cite{2019_CVPR_MAR} & CVPR'19 &67.1&48.0& 67.7&40.0\\
UCDA-CCE \cite{2019_ICCV_ucda} & ICCV'19 &55.4 & 36.7 & 64.3 & 34.5 \\
PDA-Net \cite{2019_ICCV_PDA} & ICCV'19 & 63.2 & 45.1 & \textbf{75.2} & 47.6 \\
\hline
DeepCluster \cite{2018_ECCV_deep-clustering} & ECCV'18 & 40.2&26.7 & 48.0 &26.1 \\
ours& This work &\textbf{72.1} &\textbf{53.8}&74.0&\textbf{47.9}  \\
\hline
\end{tabular}
\vspace{-0.2cm}
\end{table}

\begin{table}[t]
\centering
\scriptsize
\caption{\label{tb:stoa_PIFR}
Comparison to the state-of-the-art \textbf{supervised} PIFR models on the Multi-PIE dataset.
}
\begin{tabular}{c|c|ccccc}
\hline
Methods & avg & 0\degree & $\pm$15\degree & $\pm$30\degree & $\pm$45\degree & $\pm$60\degree \\
\hline
FIP \cite{2013_ICCV_FIP}  & 72.9& 94.3 & 90.7 & 80.7 & 64.1 & 45.9 \\
MVP \cite{2014_NIPS_MVP}& 79.3 & 95.7 & 92.8 & 83.7 & 72.9 & 60.1  \\
CPI \cite{2015_CVPR_CPI} & 83.3& \textbf{99.5} & 95.0 & 88.5 & 79.9 & 61.9  \\
DRGAN \cite{2017_CVPR_DRGAN}  & 90.2& 97.0 & 94.0 & 90.1 & 86.2 & 83.2 \\
FFGAN \cite{2017_ICCV_FFGAN} & 91.6& 95.7 & 94.6 & 92.5 & 89.7 & 85.2  \\
p-CNN \cite{2018_TIP_multitask} & 93.5& 95.4 &95.2 & 94.3 & 93.0 & 90.3  \\
\hline
DeepCluster \cite{2018_ECCV_deep-clustering} &86.4&96.7&96.6&93.3&84.8&65.6\\
 ours &\textbf{97.1}&99.1&\textbf{98.9}&\textbf{98.3}&\textbf{96.8}&\textbf{93.1} \\
\hline
\end{tabular}
\vspace{-0.2cm}
\end{table}

\begin{table}[t]
\centering
\scriptsize
\caption{\label{tb:sota_CFP}
Comparison to the state-of-the-art \textbf{supervised} PIFR models on the CFP dataset.
Format: mean(standard deviation).
}
\begin{tabular}{c|ccc}
\hline
Methods & Accuracy & EER & AUC \\
\hline
Deep Features \cite{CFP} & 84.91(1.82)&14.97(1.98)&93.00(1.55)\\
Triplet Embedding \cite{sankaranarayanan2016triplet} &89.17(2.35)&8.85(0.99)&97.00(0.53)\\
Chen et.al. \cite{chen2013blessing} &91.97(1.70)&8.00(1.68)&97.70(0.82)\\
PIM \cite{zhao2018towards}&93.10(1.01)&7.69(1.29)&97.65(0.62)\\
DRGAN \cite{2017_CVPR_DRGAN} &93.41(1.17)&6.45(0.16)&97.96(0.06)\\
p-CNN \cite{2018_TIP_multitask} &94.39(1.17)&5.94(0.11)&98.36(0.05)\\
Human &94.57(1.10)&5.02(1.07)&\textbf{98.92}(0.46)\\
\hline
DeepCluster \cite{2018_ECCV_deep-clustering}&91.30(1.58)&8.86(1.70)&96.77(0.96) \\
ours &\textbf{95.49}(0.70)&\textbf{4.74}(0.72)&98.83(0.29)\\
\hline
\end{tabular}
\vspace{-0.2cm}
\end{table}

\subsection{Comparison to the state-of-the-art methods}
\K{We further demonstrated the effectiveness of our model by comparing to the state-of-the-art methods
on both tasks}.
It should be pointed out that for RE-ID and PIFR where the goal was to solve the real-world problem,
there were no standards on architecture: different works used different networks and different pretraining data.
Thus we simply kept using the standard ResNet-50 without task-specific improvements and using the public pretraining data.
For a fairer comparison, we also compared \ws{our method with a} recent
unsupervised deep learning method DeepCluster \cite{2018_ECCV_deep-clustering},
which also \ws{uses} a discriminative classification loss.
We used the same architecture and pretraining as for our method.
%We observed that DeepCluster converged using this setting.
%To enable DeepCluster to benefit from pretraining, we dropped the Sobel filter which was originally used in  \cite{2018_ECCV_deep-clustering}
%to facilitate training from scratch.
%We kept the re-sampling strategy \cite{2018_ECCV_deep-clustering}.
% Since it was reported in \cite{2018_ECCV_deep-clustering} that over-segmentation was beneficial,
% we also set $K=2000$ and $K=500$ for DeepCluster in RE-ID and PIFR, respectively.
%We tried several cluster-updating frequency (per 100/200/400/800/1600 iterations) and reported the best results for DeepCluster  \cite{2018_ECCV_deep-clustering}.
We show the results in Table \ref{tb:stoa_REID}, \ref{tb:stoa_PIFR} and \ref{tb:sota_CFP}.

\vspace{0.1cm}
\noindent
\textbf{Superior performance across tasks and benchmarks}.
Compared to the reported results,
our method could achieve the state-of-the-art performances.
On unsupervised RE-ID task, 
our method achieved a 5.0\%/5.8\% absolute improvement in rank-1 accuracy/MAP on DukeMTMC-reID,
compared to the recent state-of-the-art RE-ID model MAR \cite{2019_CVPR_MAR},
which used exactly the same architecture and pretraining data as ours.
Although a few recent domain adaptation methods \cite{2019_ICCV_PDA} achieve
comparable performances to our method,
it is worth noting that they rely on labeled source data for discriminative learning,
while we do \emph{not} use labeled data and our method can generalize to different tasks
instead of specifically modeling a single RE-ID task.
We note that many of the compared recent state-of-the-art RE-ID methods also exploited the camera view labels \cite{2018_ECCV_HHL,2019_CVPR_invariance,2019_CVPR_MAR,2017_ICCV_asymmetric,2019_TPAMI_DECAMEL,2019_ICCV_Wu,2019_ICCV_ucda}.
For instance,
the domain adaptation RE-ID models HHL \cite{2018_ECCV_HHL}/ECN \cite{2019_CVPR_invariance}
leveraged the camera view labels to synthesize cross-view person images for training data augmentation \cite{2018_ECCV_HHL,2019_CVPR_invariance},
and MAR \cite{2019_CVPR_MAR} used view labels to learn the view-invariant soft multilabels.
\K{On the pose-invariant face recognition task, our model outperformed the state-of-the-art supervised results on both Multi-PIE and the CFP benchmarks.}
We also note that most compared PIFR models exploited both the identity labels and the pose labels.

Our model also outperformed the DeepCluster \cite{2018_ECCV_deep-clustering} significantly on all the four benchmarks.
A major reason should be that some discriminative visual clues (e.g. fine-grained clothes pattern)
of persons (/faces) were ``overpowered'' by the camera view (/pose) induced feature distortion.
Without appropriate mechanisms to address this problem,
the clustering might be misled by the feature distortion to produce inferior cluster separations.
In contrast, our model addressed this problem via
the weakly supervised decision boundary rectification and the feature drift regularization.

\begin{figure}[t]
\begin{center}
\subfigure[RE-ID, basic model]{
\includegraphics[width=0.45\linewidth]{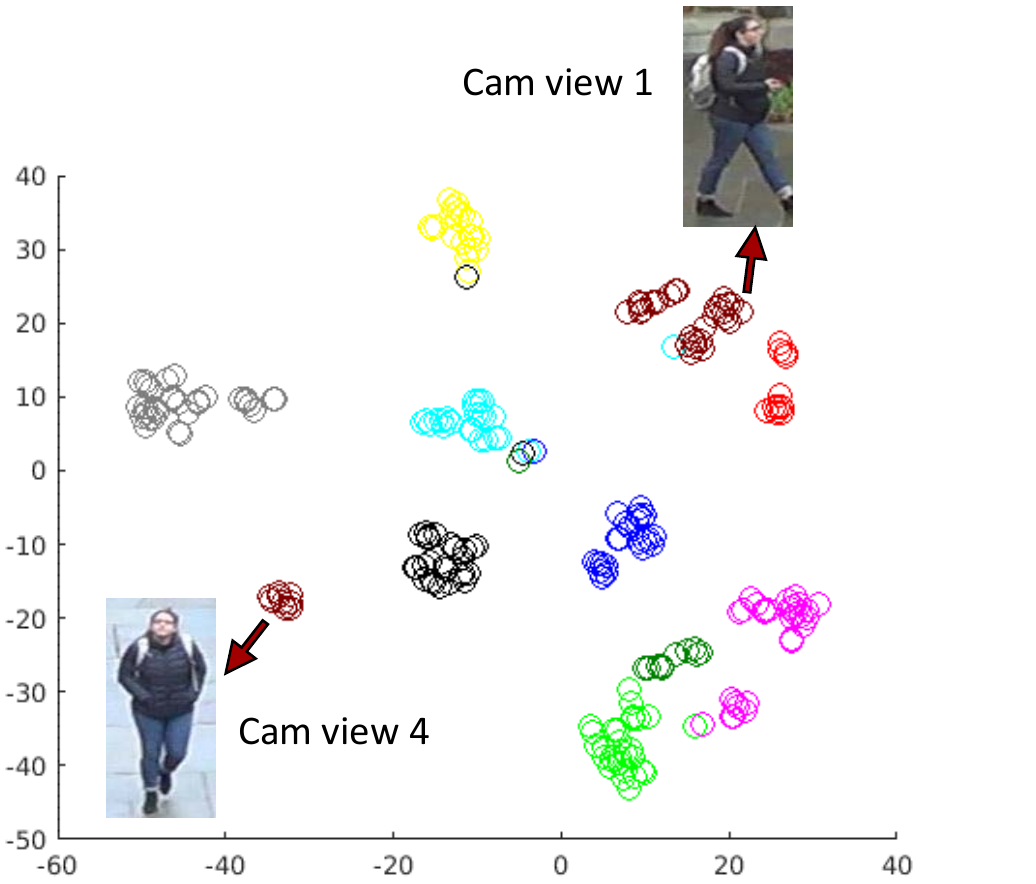}\label{fig:qualitative_reid_wo}
}
\vspace{-0.2cm}
\subfigure[RE-ID, full model]{
\includegraphics[width=0.45\linewidth]{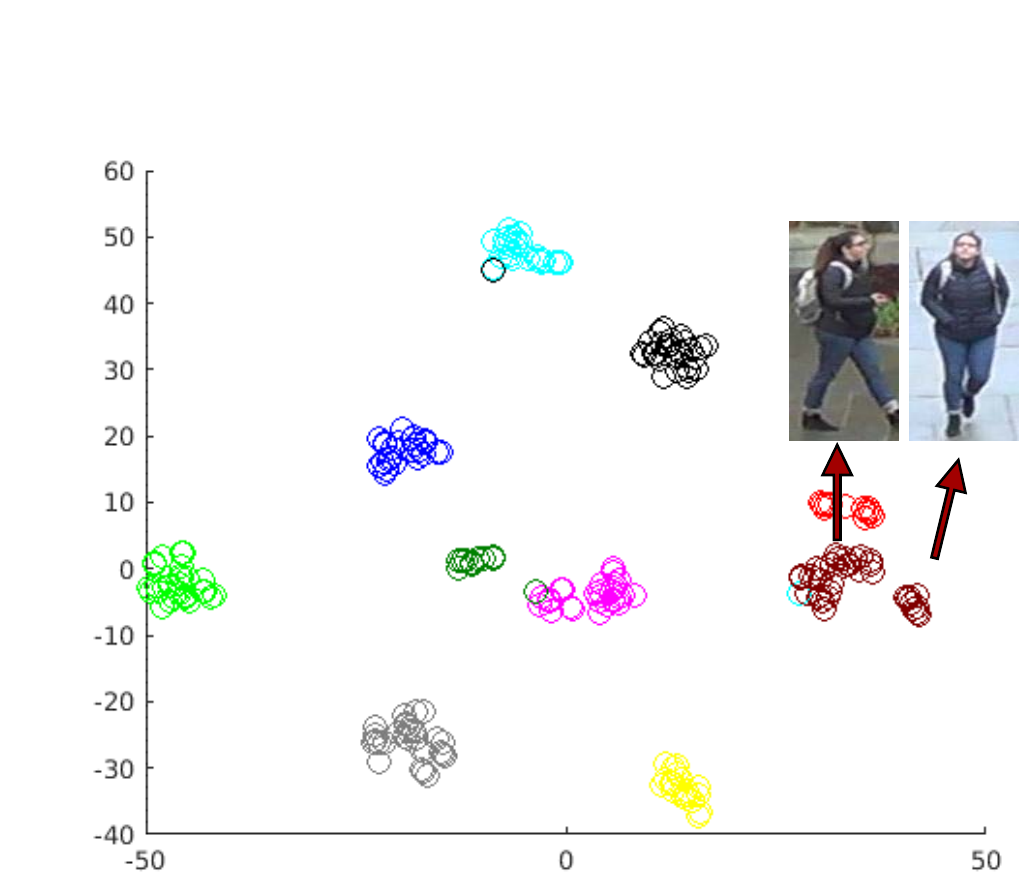}\label{fig:qualitative_reid_w}
}
\vspace{-0.2cm}
\subfigure[PIFR, basic model]{
\includegraphics[width=0.45\linewidth]{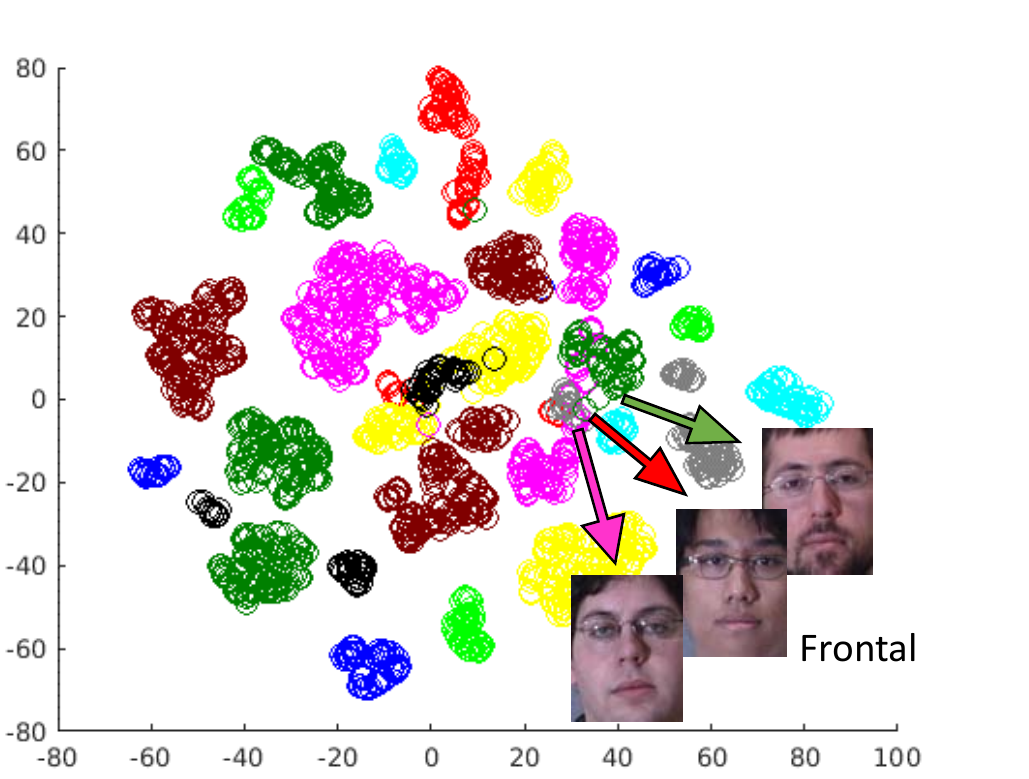}\label{fig:qualitative_face_wo}
}
\subfigure[PIFR, full model]{
\includegraphics[width=0.45\linewidth]{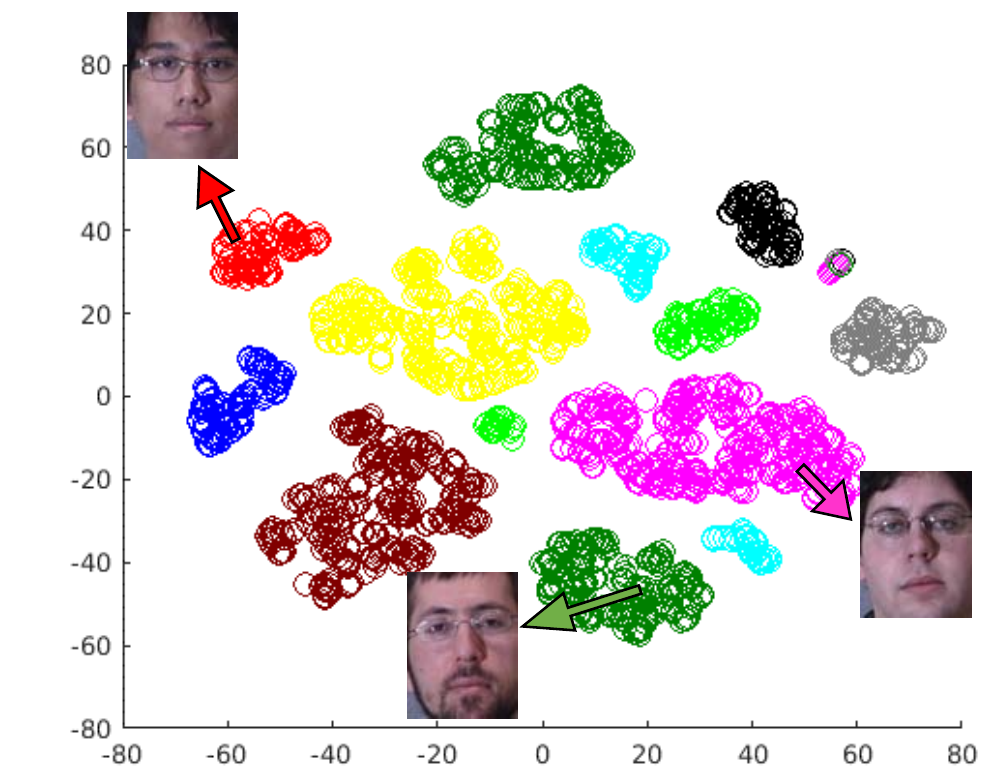}\label{fig:qualitative_face_w}
}
\end{center}
   \caption{The t-SNE embedding of 10 identities on the DukeMTMC-reID and Multi-PIE datasets.
   Each color represents an identity randomly chosen from the unseen \emph{testing} set.
   Best viewed in screen and zoom in.
   }
\label{fig:qualitative}
\vspace{-0.2cm}
\end{figure}

\vspace{-0.2cm}
\subsection{Visualization}
To provide visual insights of the problems our model tried to address,
we show the t-SNE embedding \cite{tSNE} of the learned features in Figure \ref{fig:qualitative}.
Note that the shown identities were \emph{unseen} during training,
and thus the characteristic reflected in the qualitative results were generalisable.

\vspace{0.1cm}
\noindent
\textbf{Addressing intra-identity visual feature discrepancy}.
Let us compare Figure \ref{fig:qualitative_reid_wo} and Figure \ref{fig:qualitative_reid_w} for an illustration.
Figure \ref{fig:qualitative_reid_wo} illustrated that the same person (see the highlighted brown points)
appeared differently in two camera views, which had different viewpoints and backgrounds.
This visual discrepancy caused significant feature distortion of the identity.
Apparently, it was extremely difficult to address this problem if other effective mechanisms were not provided
besides feature similarity.
From Figure \ref{fig:qualitative_reid_w} we observed that when equiped with the WDBR and WFDR,
the feature distortion was significantly alleviated.
This observation indicated that the our model leveraged the state information 
to effectively alleviate the intra-identity visual discrepancy for better discriminative feature learning.

\vspace{0.1cm}
\noindent
\textbf{Addressing inter-identity visual feature entanglement}.
In a more complex case shown in
Figure \ref{fig:qualitative_face_wo}, we observed that some visually similar frontal face images
(males with eye glasses) were entangled with each other
in the feature space learned by the basic model.
In particular, some magenta, red and dark green points highly overlapped with each other.
This demonstrated that if we simply used feature similarity, 
it was also extremely difficult to address the inter-identity visual feature entanglement.
Nevertheless, as shown in Figure \ref{fig:qualitative_face_w} our full model
could address this problem with the WDBR and WFDR.
The learned feature space was much desirable,
and the inter-identity overlapping points were now distant from each other.
In other words, our model could leverage the state information to help the unsupervised learning via alleviating the inter-identity visual feature entanglement.

\begin{table}[t]
\centering
\scriptsize
%\vspace{-0.2cm}
\caption{\label{tb:multiple}
Multiple kinds of state information on Multi-PIE. We report mean results with 5 runs (\%).
}
\begin{tabular}{c|c|ccccc}
\hline
Methods & avg & 0\degree & $\pm$15\degree & $\pm$30\degree & $\pm$45\degree & $\pm$60\degree \\
\hline
Supervised fine-tune &96.6&98.9&98.6&97.6&96.0&93.1\\
\hline
Pretrained  &83.8&95.6&94.2&89.3&81.5&64.6\\
w/ only pose labels & 94.6&96.5&96.2&95.9&94.9&90.6 \\
w/ only illumination labels &30.3&73.3&65.8&24.9&7.0&2.2\\
w/ only expression labels &43.5&80.6&75.2&51.5&26.2&2.6\\
\hline
w/ all three kinds of labels &\textbf{95.9}&\textbf{98.3}&\textbf{98.2}&\textbf{97.2}&\textbf{95.7}&\textbf{91.1} \\
\hline
\end{tabular}
\vspace{-0.4cm}
\end{table}

\subsection{Multiple kinds of state information}

Our method is easy to extend to incorporate multiple kinds of state information.
We experimented on Multi-PIE with the state information being expression, illumination and pose.
We used all 6 expressions, 20 illuminations and 9 poses.
We decomposed the rectifier by $p(k)=p_p(k)\cdot p_i(k)\cdot p_e(k)$, where the subscripts $p$/$i$/$e$ stand for pose/illumination/expression, respectively.
\kk{We also accordingly use three equally-weighted feature drift regularization terms in the loss function.}
We used hard WDBR to have a regular shape of rectifier function.
We show the results in Table \ref{tb:multiple}.
Exploiting pose labels produced much better results than illumination and expression,
indicating that pose was the most distractive on Multi-PIE.
Exploiting all three kinds of state information further improved the performance to 95.9\%,
which was closed to the supervised result 96.6\%.
This comparison showed that our model could be further improved when more valuable state information was available.

\vspace{-0.2cm}
\section{Conclusion and Discussion}
\Koven{
In this work we proposed a novel psuedo label method with state information.
We found that some proper state information could help address the visual discrepancy caused by those distractive states.
Specifically,
we investigate the state information in person re-identification and face recognition
and found the camera view labels and pose labels to be effective.
Our results indicate that it is reasonable to make use of the free state information in unsupervised person re-identification and face recognition.
Since the weakly supervised feature drift regularization (WFDR) is a simple loss term which is model-free, it can be plugged into other different methods than our proposed pseudo label method.

However, we should point out that our method works with the state information that corresponds to the visually distractive states. 
As for more general state information,
it still remains an open problem to effectively utilize it.
}

{\small
\bibliographystyle{ieee_fullname}
\bibliography{Koven}
}

\clearpage
\appendix

\section{Algorithm description and details}
We summarize our model in Algorithm \ref{alg:sasi}.
The equations in Algorithm \ref{alg:sasi} are defined in the main manuscript.

We used standard data augmentation (random crop and horizontal flip) during training.
We used the spherical feature embedding \cite{2017_ACMMM_normface,2017_CVPR_sphereface},
i.e., we enforced $||x||_2=1$ and $||\mu_k||_2=1, \forall k$.
To address the gradient saturation in the spherical embedding \cite{2017_ACMMM_normface},
we followed the method introduced by \cite{2017_ACMMM_normface} to scale every inner product $x^\mathrm{T}\mu$ in Eq. (\ref{eq:loss_aa})
up to 30.
We updated $\{p(k)\}_{k=1}^K$ every $T=40$ iterations, as we found it not sensitive in a broad range.
We maintained a buffer for $m$ and $\sigma$ as a reference,
whereas $m_j$ and $\sigma_j$ were estimated within each batch to obtain the gradient.
We updated the buffer with a momentum $\alpha=B/N$ for each batch where $B$ denoted the batch size and $N$ denotes the training set size.

\section{Probablistic interpretation of the WDBR}

In this section we first show that the weakly supervised decision boundary rectification (WDBR)
is the maximum a posteriori (MAP) optimal estimation of the surrogate label $\hat{y}$ under suitable assumptions.
Specifically, if we assume that a surrogate class is modeled by a normal distribution  \cite{2019_CVPR_MAR,2018_TPAMI_WCNN,2017_Arxiv_BEGAN} parameterized by a mean vector $\mu$ and covariance matrix $\bm{\Sigma}$,
the likelihood that $x$ is generated by the $y$-th surrogate class is:
\begin{align}\label{eq:likelihood}
p(x|y) = \frac{\exp(-||x-\mu_y||^2/2)}{\Sigma_{k=1}^K \exp(-||x-\mu_k||^2/2)},
\end{align}
where we assume the identity covariance matrix \cite{2019_CVPR_MAR,2018_TPAMI_WCNN,2017_Arxiv_BEGAN}, i.e. $\bm{\Sigma}_k = I, \forall k$.
Since we enforce $||x||_2=1$ and $||\mu_k||_2=1, \forall k$,
we have $-||x-\mu_{y}||^2/2=x^\mathrm{T}\mu_{y}+1$.
Then Eq. (\ref{eq:likelihood}) is equivalent to:
\begin{align}\label{eq:likelihood2}
p(x|y) = \frac{\exp(x^\mathrm{T}\mu_{y})}{\Sigma_{k=1}^K \exp(x^\mathrm{T}\mu_{k})}.
\end{align}
From Eq. (\ref{eq:likelihood2}) we can see that the basic surrogate classification in Eq. (1) in the main manuscript is the Maximum Likelihood Estimation (MLE) of model parameters $\theta$ and $\{\mu_k\}$.
And the assignment
\begin{align}\label{eq:assignment}
\hat{y} = \arg\max_k \: \exp(x^\mathrm{T}\mu_{k}) = \arg\max_k \: p(x| k)
\end{align}
is the MLE optimal assignment.
If we further consider the prior information of each surrogate class,
i.e., which surrogate classes are more preferable to assign,
we can improve the assignment to the Maximum a Posteriori (MAP) optimal assignment:
\begin{align}\label{eq:bayesian}
\hat{y} = \arg\max_k \: p(k)\exp(x^\mathrm{T}\mu_{k}) = \arg\max_k \: p(k| x),
\end{align}
where
\begin{align}\label{eq:bayesian2}
p(k| x) = \frac{p(y)\exp(x^\mathrm{T}\mu_{y})}{\Sigma_{k=1}^K p(k)\exp(x^\mathrm{T}\mu_{k})}
\end{align}
is the posterior probability.
Eq. (\ref{eq:bayesian}) is identical to the rectified assignment in Eq. (6) in the main manuscript.
Hence, we can interpret the weakly supervised rectifier function as a prior probability
that specifies our preference on the surrogate class.
In particular, when we use the hard rectification,
we actually specify that we dislike severely unbalanced surrogate classes.
When we use the soft rectification,
we specify that we favor the more balanced surrogate classes.

\vspace{0.1cm}
\noindent
\textbf{Derivation of the decision boundary}.
Here we consider the simplest two-surrogate class case. 
It is straightforward to extend it to multi-surrogate class cases.
From Eq. (\ref{eq:bayesian}) we can see
that the decision boundary between two surrogate class $\mu_1$ and $\mu_2$ is:
\begin{align}
&p(1)\exp(x^\mathrm{T}\mu_{1}) = p(2)\exp(x^\mathrm{T}\mu_{2}) \\ \nonumber
\Rightarrow \quad &\exp(x^\mathrm{T}\mu_1+\log p(1)) = \exp(x^\mathrm{T}\mu_2+\log p(2)) \\ \nonumber
\Rightarrow \quad &(\mu_1-\mu_2)^\mathrm{T}x + \log p(1) - \log p(2) = 0 \\ \nonumber
\Rightarrow \quad &(\mu_1-\mu_2)^\mathrm{T}x +\log\frac{p(1)}{p(2)} = 0.
\end{align}

\begin{algorithm}[t]\label{alg:sasi}
	\footnotesize
	\caption{Weakly supervised discriminative learning}
	%\SetKwInOut{Input}{Input}
	%\SetKwInOut{Output}{Output}
	\textbf{Input}: Training set $\mathcal{U}=\{u_i\}$, state information $\mathcal{S}=\{s_i\}$, pretrained model $f(\cdot,\theta^{(0)})$\\
	\textbf{Output}: Learned model $f(\cdot,\theta)$\\
	\textbf{Initialization}: \\
	Obtain the initial feature space $\mathcal{X}_{init}$ = $f(\mathcal{U},\theta^{(0)})$.\\
	Initialize surrogate classifiers $\{\mu_k\}_{k=1}^K$ as the centroids obtained by performing standard K-means clustering
	on $\mathcal{X}_{init}$. \\
	Initialize the surrogate rectifiers $p(k)=1, k=1,\cdots, K$.\\
	Initialize total distribution vectors $m$/$\sigma$ on $\mathcal{X}_{init}$.\\
	\textbf{Training}:\\
	\For{the $i$-th batch $\{\mathcal{U}^{(i)}, \mathcal{S}^{(i)}\}$}
	{
		Obtain the features in the batch $\mathcal{X}^{(i)} = f(\mathcal{U}^{(i)}, \theta)$.\\
		Assign every $x\in \mathcal{X}^{(i)}$ to a surrogate class $\hat{y}$ by Eq. (6) in the main manuscript.\\
		Estimate the state sub-distributions  $\{m_j,\sigma_j\}_{j=1}^J$ in this batch.\\
		Compute the loss (Eq. (11) in the main manuscript) and update the model.\\
		Estimate the total distribution in this batch: $m^{(i)}$/$\sigma^{(i)}$.\\
		Update $m$ by $m \leftarrow (1-\alpha)m+\alpha m^{(i)}$ and $\sigma$ by $\sigma \leftarrow (1-\alpha)\sigma+\alpha\sigma^{(i)}$.\\
		Obtain $\{R_k\}_{k=1}^K$ by Eq. (3) and update $\{p(k)\}_{k=1}^K$ by Eq. (7) every $T$ batches.\\
	}
	\textbf{Testing}:\\
	Discard $\{\mu_k\}_{k=1}^K$ and use $f(\cdot|\theta)$ to extract the discriminative feature.
\end{algorithm}

\section{Hyperparameter evaluation}\label{sec:hyper}

In order to provide insight and guidance on choosing the hyperparameter values,
%which is very important in unsupervised learning due to the lack of standard techniques like cross-validation,
in this section we show evaluation results of the hyperparameters to reveal some behaviors and characteristics of our model.
For person re-identification (RE-ID) we evaluated on the widely-used Market-1501 and DukeMTMC-reID datasets,
and for pose-invariant face recognition (PIFR) we evaluated on the large-scale Multi-PIE dataset.
For easier interpretation and more in-depth analysis, we used the hard rectification function
on \emph{all} datasets.
This was because the hard rectification function could be interpreted as nullification of high maximum predominance index (and thus likely
to be dominated by the extrinsic state) surrogate classes.

\begin{figure}[t]
	\begin{center}
		\subfigure[RE-ID]{
			\includegraphics[width=0.4\linewidth,height=0.28\linewidth]{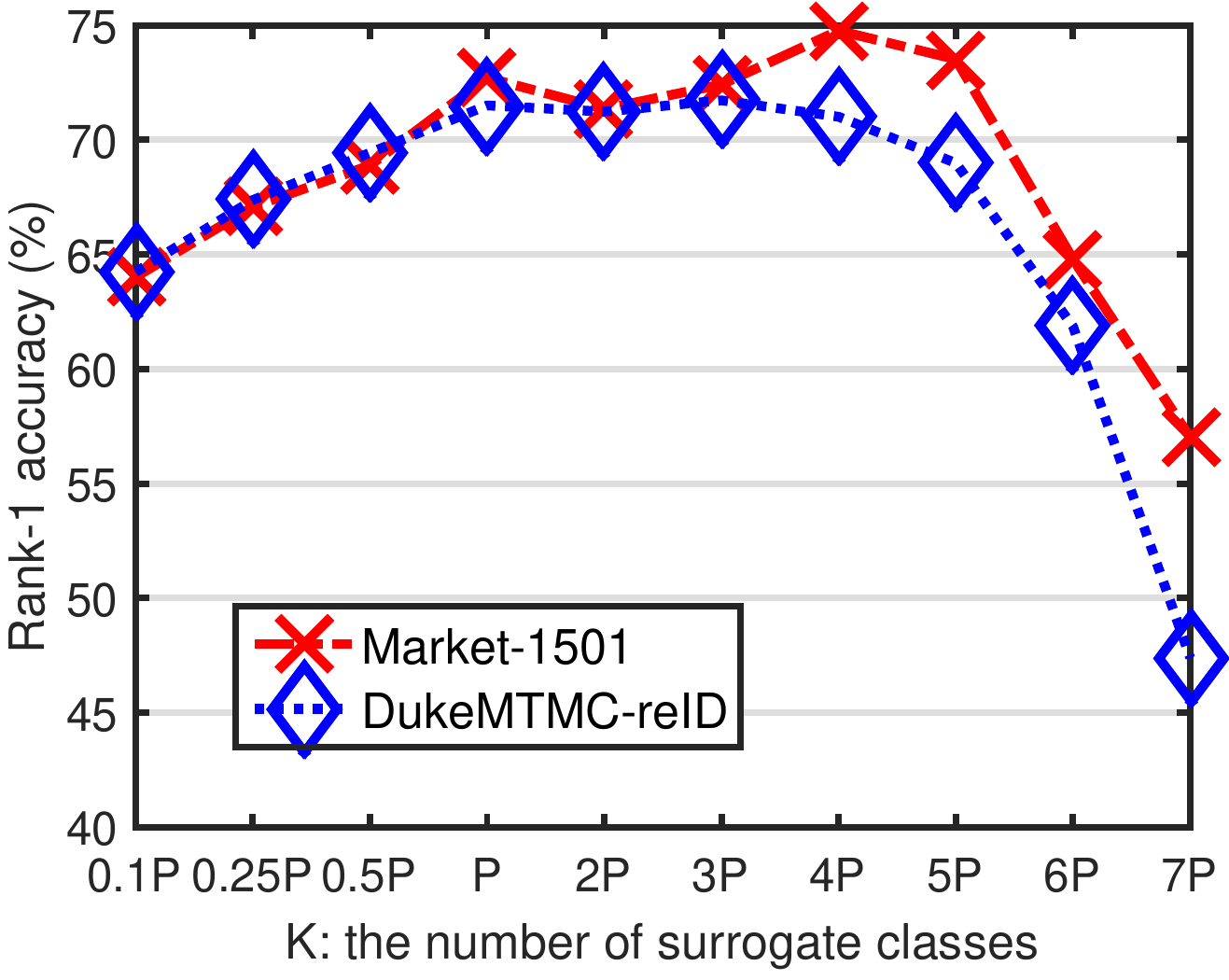}\label{fig:K_vary_reid}
		}
		\vspace{-0.2cm}
		\subfigure[PIFR]{
			\includegraphics[width=0.4\linewidth,height=0.28\linewidth]{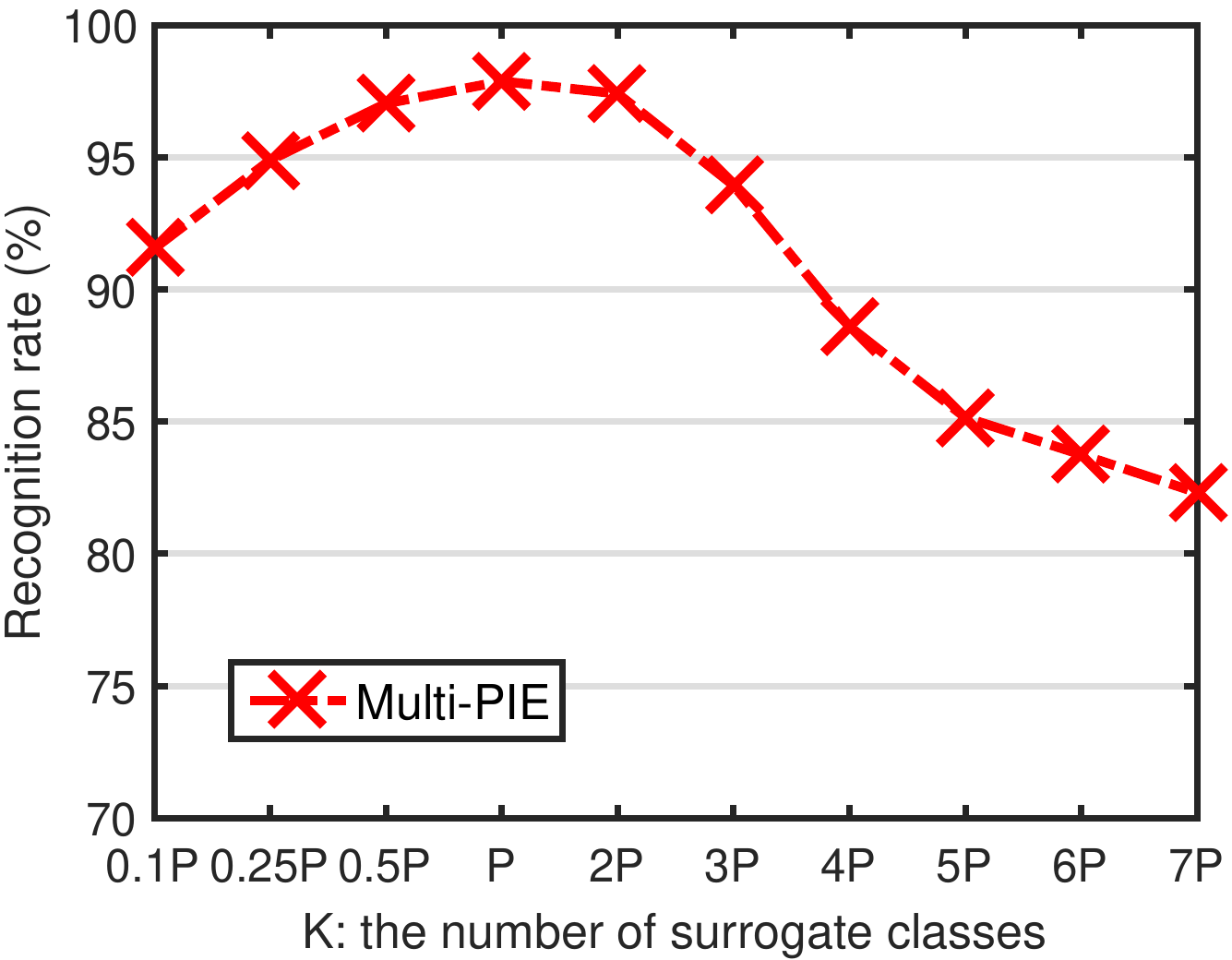}\label{fig:K_vary_face}
		}
		\vspace{-0.2cm}
	\end{center}
	\caption{Evaluation for the number of surrogate classes $K$.
		$P$ is the precise number of classes in the training set.
	}
	\vspace{-0.2cm}
	\label{fig:K_vary}
\end{figure}

\vspace{0.1cm}
\noindent
\textbf{$K$: Number of surrogate classes}.
In each task (i.e. RE-ID or PIFR),
we varied the number of surrogate classes $K$
by setting it to a multiple of the precise number of classes in the training set $P$ (e.g. $P=750$ for Market-1501 and $P=200$ for Multi-PIE).
We show the results in Figure \ref{fig:K_vary}.
From Figure S1(a) and S1(b) we can see that
the optimal performances could be achieved when $K=P$ or $K>P$.
This might be because the dynamic nullification (i.e. hard rectification) reduced the effective $K$ in training,
so that a larger $K$ could also be optimal.
In a practical perspective, we might estimate an ``upper bound'' of $P$ and set $K$ to it according to some prior knowledge.

% As $K$ became too low like $K=0.1P$ or too high like $K=7P$,
% the performances dropped noticeably.
% However, we observed that the performance was consistently stable in the same range
% (e.g. $K\in [0.25P, 5P]$) in different RE-ID datasets.
% Furthermore, the same pattern was observed in the PIFR task shown in Figure \ref{fig:K_vary_face}.

\begin{figure}[t]
	\begin{center}
		\subfigure[RE-ID]{
			\includegraphics[width=0.42\linewidth,height=0.28\linewidth]{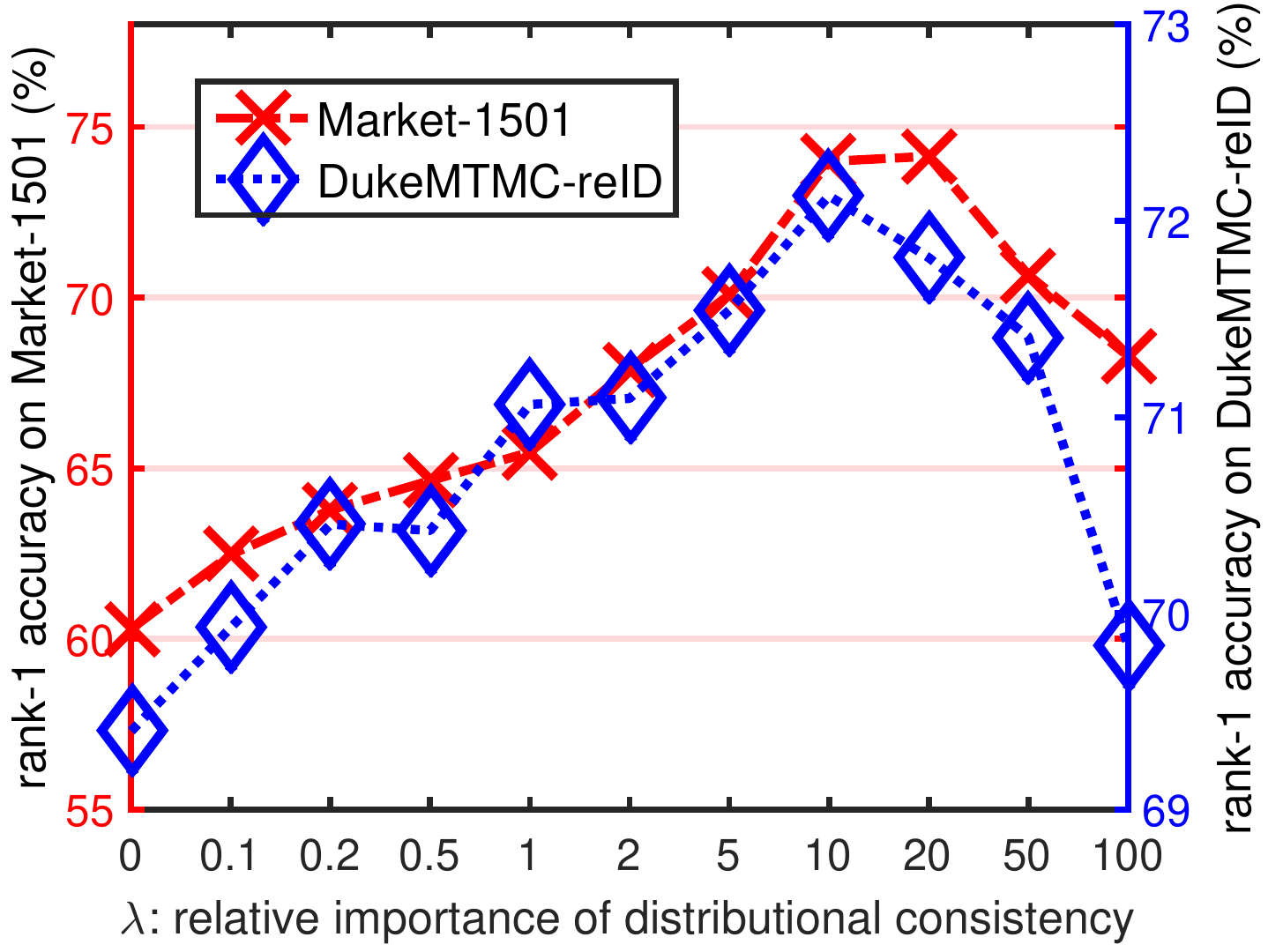}\label{fig:lambda_vary_reid}
		}
		\vspace{-0.2cm}
		\subfigure[PIFR]{
			\includegraphics[width=0.4\linewidth,height=0.28\linewidth]{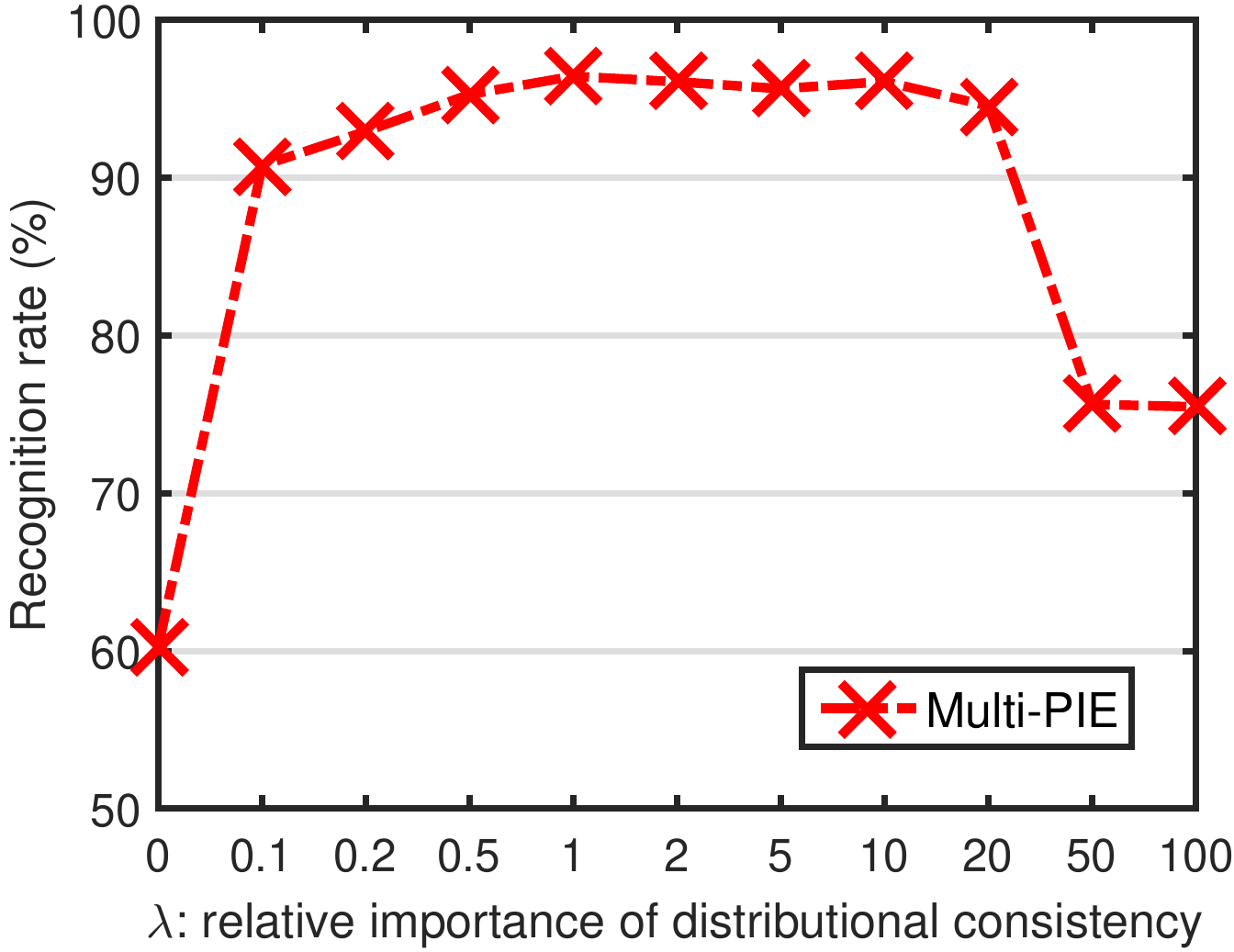}\label{fig:lambda_vary_face}
		}
		\vspace{-0.2cm}
	\end{center}
	\caption{Evaluation for the feature drift regularization weight $\lambda$.
	}
	\vspace{-0.2cm}
	\label{fig:lambda_vary}
\end{figure}

\vspace{0.1cm}
\noindent
\textbf{$\lambda$: Weight of feature drift regularization}.
We show the evaluation results in Figure \ref{fig:lambda_vary}.
Here we removed the surrogate decision boundary rectification for PIFR
to better understand the characteristic of $\lambda$.
From Figure S2(a) and S2(b),
we found that while the performances on RE-ID were optimal around $\lambda=10$,
for PIFR it was near optimal within the range of $[0.5,20]$.

\begin{figure}[t]
	\begin{center}
		\includegraphics[width=0.6\linewidth]{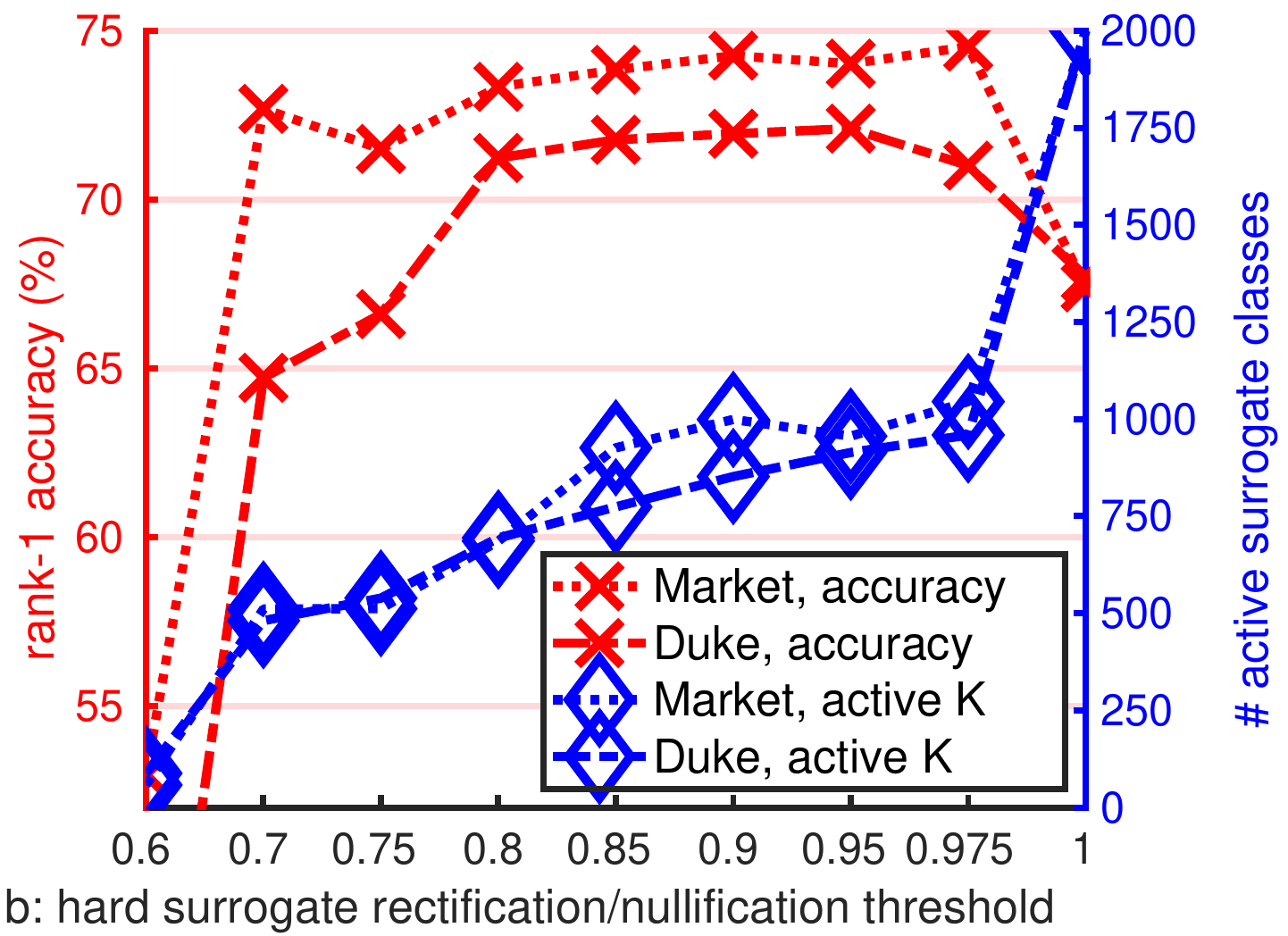}
		\vspace{-0.2cm}
	\end{center}
	\caption{Evaluation for the rectification threshold $b$ (hard threshold here).
		We also show the active number of surrogate classes in the final convergence epoch (denoted as ``active K'' in the legend).
	}
	\vspace{-0.2cm}
	\label{fig:gamma_vary}
\end{figure}

\vspace{0.1cm}
\noindent
\textbf{Hard surrogate rectification/nullification threshold}.
We show the evaluation results on RE-ID datasets in Figure \ref{fig:gamma_vary}.
The performances were optimal when $b$ was not too low, e.g. $b \in [0.8,0.9]$
was optimal for both RE-ID datasets.
A major reason was that it was difficult to form sufficient surrogate classes when the threshold was too low.
To see this, in Figure \ref{fig:gamma_vary} we also show the number of active (i.e. not nullified) surrogate classes
in the final convergence epoch.
Clearly, a lack of surrogate classes was harmful to the discriminative learning.

\begin{figure}[t]
	\begin{center}
		\includegraphics[width=0.6\linewidth]{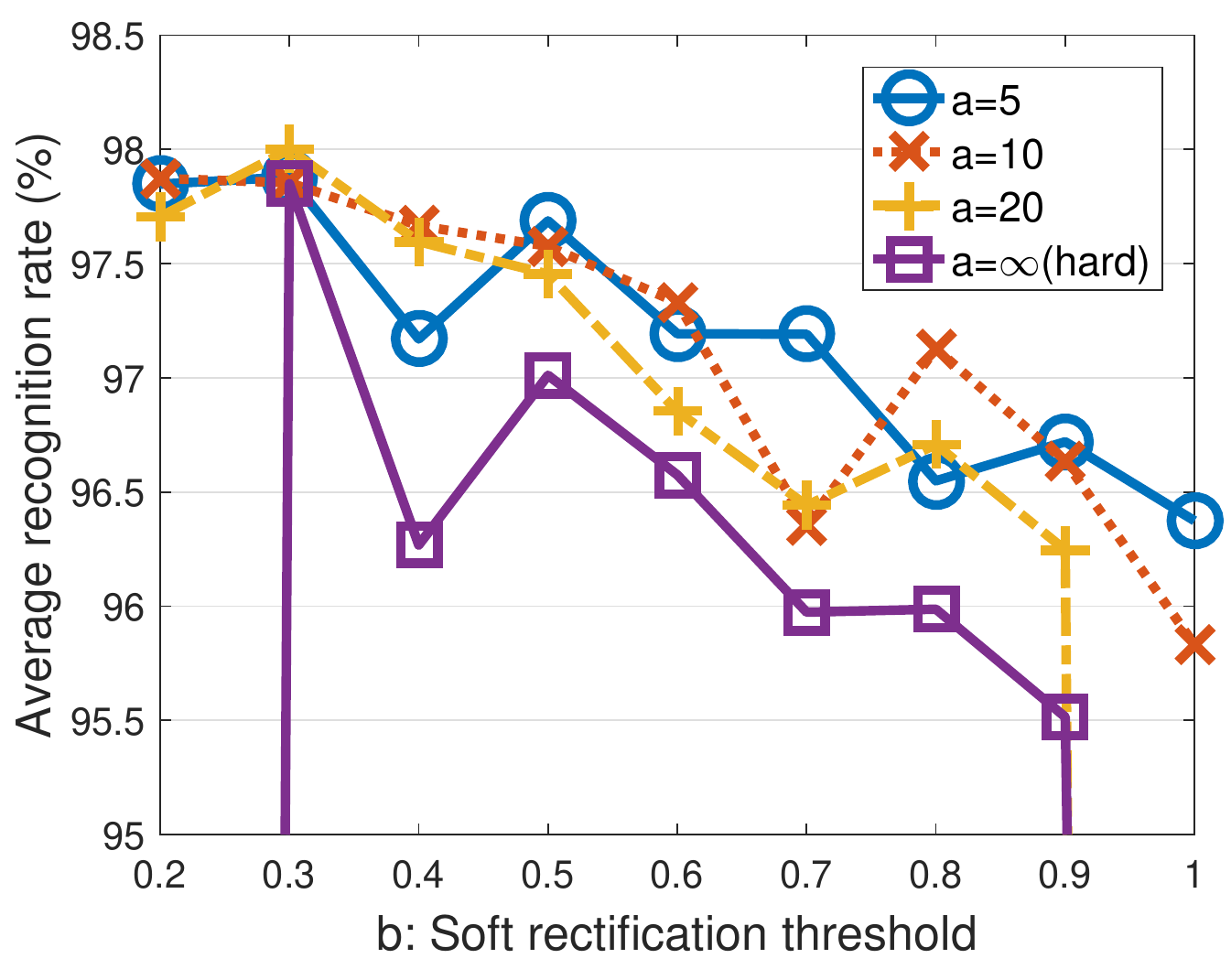}
		\vspace{-0.2cm}
	\end{center}
	\caption{Soft surrogate rectification evaluation on Multi-PIE.
		$a$ denotes the rectification strength.
	}
	\vspace{-0.2cm}
	\label{fig:soft}
\end{figure}

\vspace{0.1cm}
\noindent
\textbf{Soft surrogate rectification}.
We show the evaluation results on Multi-PIE in Figure \ref{fig:soft}.
As analyzed in the main manuscript,
soft rectification consistently improved over the hard rectification
due to the balanced classes on Multi-PIE.
The optimal value of the soft rectification threshold was around $0.3$
because on Multi-PIE the five poses were evenly distributed
and thus the optimal MPI shall be around slightly above $1/5$.
In a practical perspective,
when we have prior knowledge of the unlabelled data,
we might be able to estimate the soft rectification threshold.
Nevertheless, even when we do not have reliable prior knowledge,
the robust conservative hard rectification could also be effective.

\section{Simulating using estimated pose labels}
In this supplementary material we present the evaluation on our method's robustness for pose label perturbation.
This is a simulation for the more challenging real-world PIFR setting,
where the pose labels are obtained by pose estimation models,
and thus there might be incorrect pose labels.
We note that this is not the case for person re-identification (RE-ID),
because in RE-ID every image comes from a certain camera view of the surveillance camera network,
so that no estimation is involved.

\begin{figure}[t]
	\begin{center}
		\includegraphics[width=0.6\linewidth]{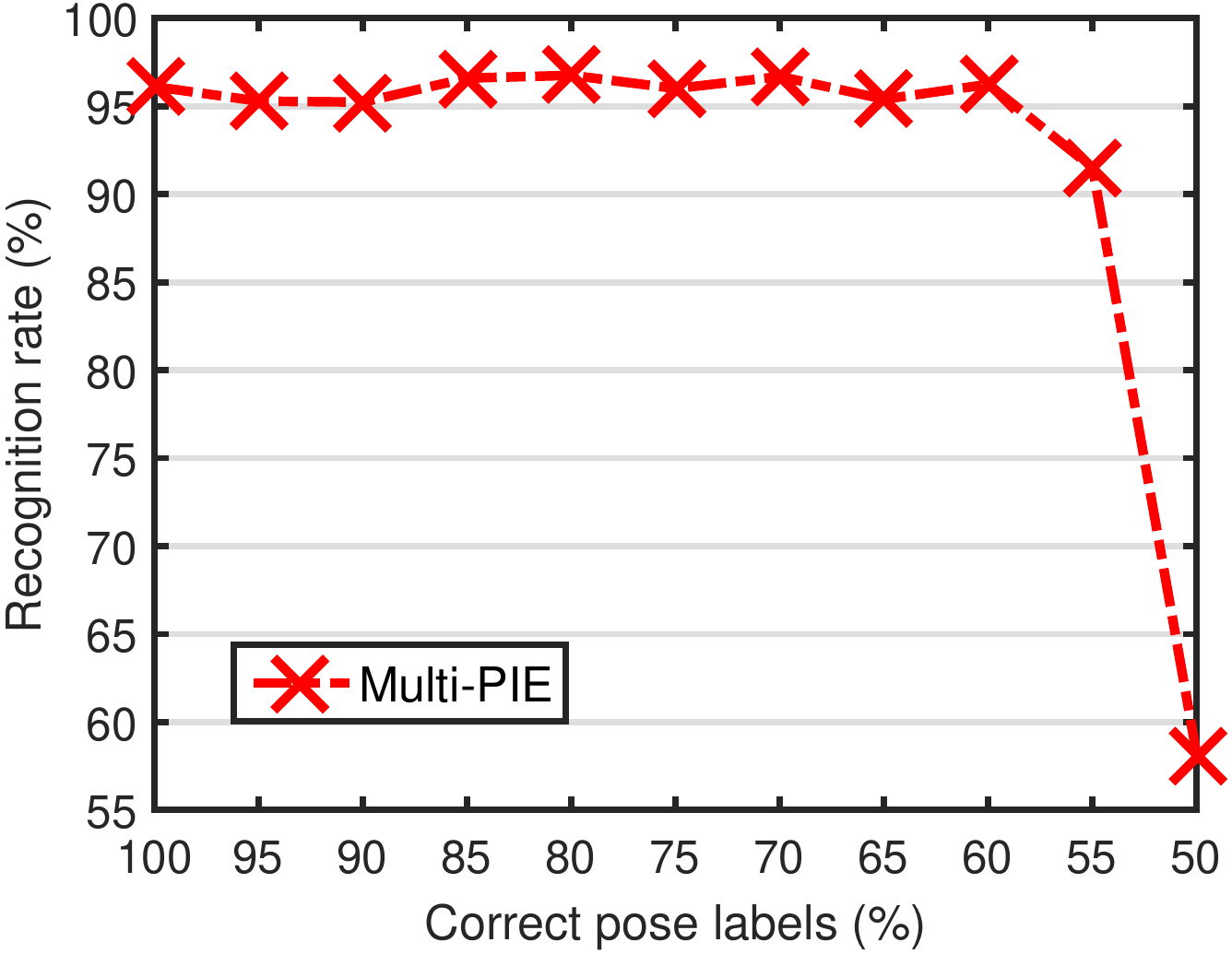}
	\end{center}
	\caption{Evaluation for noisy pose labels on Multi-PIE.
	}
	\label{fig:c_vary}
\end{figure}

To simulate the pose label noise, we add perturbation to the groundtruth pose labels.
We randomly reset some pose labels to incorrect values (e.g. we reset a randomly chosen pose label to 15\degree which is actually 60\degree).
The randomly reset pose labels were equally distributed in every degree.
For example, when we reset 20\% pose labels,
there were 20\% of 60\degree pose labels were reset incorrect, 
20\% of 45\degree pose labels were incorrect, and so forth for other degrees.
We vary the incorrect percents and show the results in Figure \ref{fig:c_vary}.

From Figure \ref{fig:c_vary} we observed that the performance on PIFR did not drop significantly until less than 60\% pose labels were correct.
This observation indicated that our model could tolerate a moderate extent of pose label noise.
A major reason was that when a few pose labels were incorrect,
a highly affected surrogate class (whose members were mostly of the same pose) 
would still have a high Maximum Predominance Index
for it to be nullfied.
In addition, when most pose labels were correct 
the estimation of the manifestation sub-distributions should
approximate the correct manifestation sub-distributions.
Therefore, our model should be robust for the unsupervised PIFR task
when a few pose labels were perturbated.

%
%We used the MTCNN \cite{MTCNN} to detect faces and produce the landmark points.
%Then we used the pose estimation toolkit provided by OpenCV\footnote{\url{https://www.learnopencv.com/head-pose-estimation-using-opencv-and-dlib/}}
%to estimate the pose labels.
%Since on Multi-PIE \cite{MPIE} the pose refers to the yaw (horizontal view angle),
%we only used the yaw angle.
%Let $y$ be the estimated yaw angle of a face image.
%We assigned the pose label $p$ to the face image by the following function:
%\begin{align}
%p =
%\begin{cases}
%-60\degree, & \text{if } y < -52.5\degree \\
%-45\degree, & \text{if } y \in [-52.5\degree, -37.5\degree) \\
%-30\degree, & \text{if } y \in [-37.5\degree, 22.5\degree) \\
%-15\degree, & \text{if } y \in [-22.5\degree, -7.5\degree) \\
%0\degree, & \text{if } y \in [-7.5\degree, 7.5\degree] \\
%15\degree, & \text{if } y \in (7.5\degree, 22.5\degree] \\
%30\degree, & \text{if } y \in (22.5\degree, 37.5\degree] \\
%45\degree, & \text{if } y \in (37.5\degree, 52.5\degree] \\
%60\degree, & \text{if } y > 52.5\degree,
%\end{cases}
%\end{align}
%Other settings were kept the same.
%The accuracy of the estimated pose labels was $?$.

%{\small
%\bibliographystyle{ieee}
%\bibliography{Koven}
%}

\end{document}